\documentclass[sigconf]{acmart}

\usepackage{pifont}
\usepackage{subcaption}
\usepackage{wasysym}
\usepackage{stfloats}
\usepackage{multirow} 
\usepackage{overpic}
\usepackage{graphicx}
\usepackage{caption}
\usepackage{xcolor}
\usepackage{url}

\colorlet{mygreen}{green!50!black}
\colorlet{myred}{red!50!black}

\AtBeginDocument{%
  \providecommand\BibTeX{{%
    \normalfont B\kern-0.5em{\scshape i\kern-0.25em b}\kern-0.8em\TeX}}}

\copyrightyear{2023}
\acmYear{2023}
\setcopyright{acmlicensed}
\acmConference[MM '23] {Proceedings of the 31st ACM International Conference on Multimedia}{October 29--November 3, 2023}{Ottawa, ON, Canada.}
\acmBooktitle{Proceedings of the 31st ACM International Conference on Multimedia (MM '23), October 29--November 3, 2023, Ottawa, ON, Canada}
\acmPrice{15.00}
\acmISBN{979-8-4007-0108-5/23/10}
\acmDOI{10.1145/3581783.3611923}
\begin{document}
\title{Light-VQA: A Multi-Dimensional Quality Assessment Model for Low-Light Video Enhancement}

\author{Yunlong Dong}
\affiliation{%
  \institution{Shanghai Jiao Tong University}
  \city{Shanghai}
  \country{China}}
\email{dongyunlong@sjtu.edu.cn}

\author{Xiaohong Liu$^*$}
\affiliation{%
  \institution{Shanghai Jiao Tong University}
  \city{Shanghai}
  \country{China}}
\email{xiaohongliu@sjtu.edu.cn}

\author{Yixuan Gao}
\affiliation{%
  \institution{Shanghai Jiao Tong University}
  \city{Shanghai}
  \country{China}}
\email{gaoyixuan@sjtu.edu.cn}

\author{Xunchu Zhou}
\affiliation{%
  \institution{Shanghai Jiao Tong University}
  \city{Shanghai}
  \country{China}}
\email{zhou_xc@sjtu.edu.cn}

\author{Tao Tan}
\affiliation{%
  \institution{Macao Polytechnic University}
  \city{Macao}
  \country{China}}
\email{taotanjs@gmail.com}

\author{Guangtao Zhai}
\affiliation{%
  \institution{Shanghai Jiao Tong University}
  \city{Shanghai}
  \country{China}}
\email{zhaiguangtao@sjtu.edu.cn}

\thanks{$^*$Corresponding authors.}


\begin{abstract}
Recently, Users Generated Content (UGC) videos becomes ubiquitous in our daily lives. 
However, due to the limitations of photographic equipments and techniques, UGC videos often contain various degradations, in which one of the most visually unfavorable effects is the underexposure. Therefore, corresponding video enhancement algorithms such as Low-Light Video Enhancement (LLVE) have been proposed to deal with the specific degradation. However, different from video enhancement algorithms, almost all existing Video Quality Assessment (VQA) models are built generally rather than specifically, which measure the quality of a video from a comprehensive perspective. To the best of our knowledge, there is no VQA model specially designed for videos enhanced by LLVE algorithms. To this end, we first construct a Low-Light Video Enhancement Quality Assessment (LLVE-QA) dataset in which 254 original low-light videos are collected and then enhanced by leveraging 8 LLVE algorithms to obtain 2,060 videos in total. 
Moreover, we propose a quality assessment model specialized in LLVE, named Light-VQA. More concretely, since the brightness and noise have the most impact on low-light enhanced VQA, we handcraft corresponding features and integrate them with deep-learning-based semantic features as the overall spatial information. As for temporal information, in addition to deep-learning-based motion features, we also investigate the handcrafted brightness consistency among video frames, and the overall temporal information is their concatenation. Subsequently, spatial and temporal information is fused to obtain the quality-aware representation of a video.
Extensive experimental results show that our Light-VQA achieves the best performance against the current State-Of-The-Art (SOTA) on LLVE-QA and public dataset. \textit{Dataset and Codes can be found at \url{https://github.com/wenzhouyidu/Light-VQA}.}
\end{abstract}



\begin{teaserfigure}
\centering
\begin{overpic}[width=0.95\textwidth]{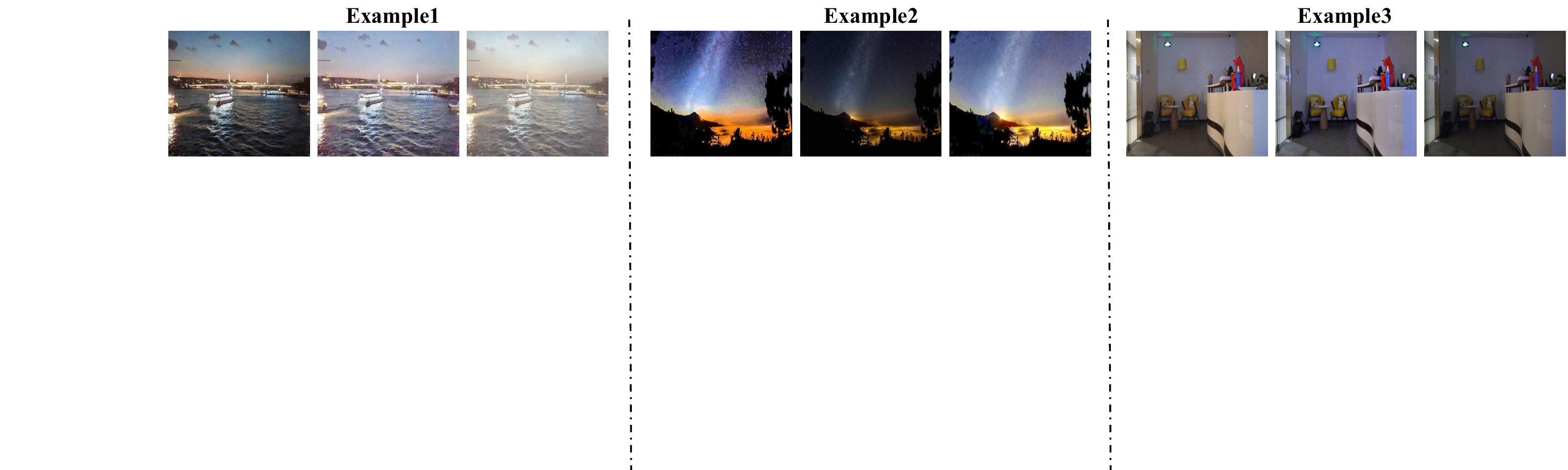}
\put(11.8, 18.5){\textbf{\scriptsize AGCCPF~\cite{gupta2016minimum}}}
\put(21, 18.5){\textbf{\scriptsize DCC-Net~\cite{zhang2022deep}}}
\put(30, 18.5){\textbf{\scriptsize StableLLVE~\cite{zhang2021learning}}}
\put(44, 18.5){\textbf{\scriptsize GHE~\cite{abutaleb1989automatic}}}
\put(51.8, 18.5){\textbf{\scriptsize BPHEME~\cite{wang2005brightness}}}
\put(61, 18.5){\textbf{\scriptsize MBLLVEN~\cite{lv2018mbllen}}}
\put(72.5, 18.5){\textbf{\scriptsize DCC-Net~\cite{zhang2022deep}}}
\put(83.2, 18.5){\textbf{\scriptsize SGZSL~\cite{zheng2022semantic}}}
\put(92.2, 18.5){\textbf{\scriptsize CapCut~\cite{capcut-web}}}
\put(-2.3, 16){V-BLIINDS~\cite{saad2014blind}}
\put(24,16){\textcolor{mygreen}{\ding{52}}}
\put(45.5,16){\textcolor{mygreen}{\ding{52}}}
\put(76,16){\textcolor{mygreen}{\ding{52}}}
\put(-2.3, 13.43){TLVQM~\cite{korhonen2019two}}
\put(24,13.43){\textcolor{mygreen}{\ding{52}}}
\put(45.5,13.43){\textcolor{mygreen}{\ding{52}}}
\put(85,13.43){\textcolor{mygreen}{\ding{52}}}
\put(-2.3, 10.86){VIDEVAL~\cite{tu2021ugc}}
\put(34,10.86){\textcolor{mygreen}{\ding{52}}}
\put(64.5,10.86){\textcolor{mygreen}{\ding{52}}}
\put(85,10.86){\textcolor{mygreen}{\ding{52}}}
\put(-2.3, 8.29){RAPIQUE~\cite{tu2021rapique}}
\put(14.7,8.29){\textcolor{mygreen}{\ding{52}}}
\put(64.5,8.29){\textcolor{mygreen}{\ding{52}}}
\put(95,8.29){\textcolor{mygreen}{\ding{52}}}
\put(-2.3, 5.72){Simple-VQA~\cite{sun2022deep}}
\put(34,5.72){\textcolor{mygreen}{\ding{52}}}
\put(45.5,5.72){\textcolor{mygreen}{\ding{52}}}
\put(76,5.72){\textcolor{mygreen}{\ding{52}}}
\put(-2.3, 3.15){FAST-VQA~\cite{wu2022fast}}
\put(34,3.15){\textcolor{mygreen}{\ding{52}}}
\put(45.5,3.15){\textcolor{mygreen}{\ding{52}}}
\put(76,3.15){\textcolor{mygreen}{\ding{52}}}
\put(-2.3, 0.57){Light-VQA (Ours)}
\put(24,0.57){\textcolor{mygreen}{\ding{52}}}
\put(64.5,0.57){\textcolor{mygreen}{\ding{52}}}
\put(76,0.57){\textcolor{mygreen}{\ding{52}}}
\put(-2.3, -2){MOS}
\put(24,-2){\textcolor{mygreen}{\ding{52}}}
\put(64.5,-2){\textcolor{mygreen}{\ding{52}}}
\put(76,-2){\textcolor{mygreen}{\ding{52}}}
\end{overpic}
\vspace{0.5em}
\caption{\textcolor{myred}{Which video has the best visual perceptual quality in each example listed?} The above 9 figures are representative frames of sample enhanced videos obtained by applying different enhancement algorithms to corresponding original low-light videos. The concrete algorithms are listed below the figures.
Then we use 6 state-of-the-art VQA models (V-BLIINDS~\cite{saad2014blind}, TLVQM~\cite{korhonen2019two}, VIDEVAL~\cite{tu2021ugc}, RAPIQUE~\cite{tu2021rapique}, Simple-VQA~\cite{sun2022deep}, and FAST-VQA~\cite{wu2022fast}) and the proposed Light-VQA to predict the quality of these enhanced videos. The check marks represent the enhanced video with the best perceptual quality predicted by each model.
Mean Opinion Scores (MOSs), the ground-truth perceptual quality of enhanced videos, are obtained through a subjective experiment.
It is evident that the prediction results of Light-VQA are highly consistent with human perception as compared to others. More detailed qualitative results can be found in Supplementary.
}
\label{fig:teaser}
\end{teaserfigure}
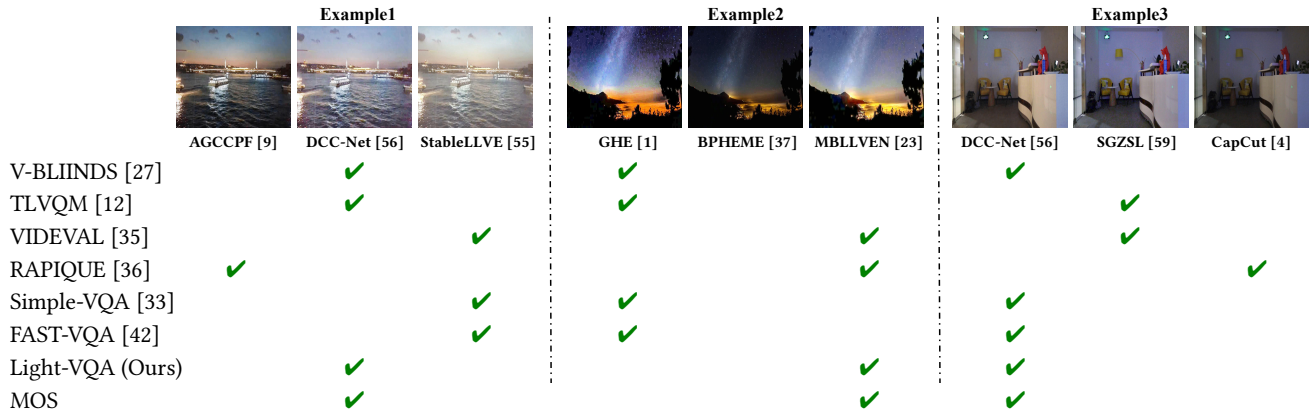


\maketitle

\section{Introduction}
Compared to text and images also spreading widely~\cite{zhang2023aigc, aigcqa}, videos are generally more entertaining and informative.
However, due to the influence in photographic devices and skills, the quality of UGC videos often varies greatly. It is frustrating that the precious and memorable moment is degraded by photographic limitations (\textit{e.g.}, underexposure, low frame-rate, and low resolution). To address the problems mentioned above, specific video enhancement algorithms have been proposed, such as Low-Light Video Enhancement (LLVE)~\cite{gupta2016minimum, wang2005brightness, zheng2022semantic, zhang2022deep}, video frame interpolation~\cite{shi2022video, shi2021video, niklaus2017video, bao2019depth}, and super-resolution reconstruction~\cite{shi2021learning, liu2021exploit, liu2020end, liu2018robust, yin2023online}.
In this paper, we focus on the quality assessment of enhanced low-light videos. Low-light videos are often captured in the low- or back-lighting environments and suffer from significant degradations such as low visibility and noises~\cite{zhai2021perceptual}. Such degraded videos will challenge many computer vision downstream tasks~\cite{zheng2020optical} such as object detection, semantic segmentation, \textit{etc.}, which are usually resorted to videos with good quality. Therefore, many LLVE algorithms have been developed to improve the visual quality of low-light videos. To this end, 
one straightforward way is to split the video into frames and apply Low-Light Image Enhancement (LLIE) algorithms to enhance each frame of this video. Representative traditional LLIE algorithms include AGCCPF~\cite{gupta2016minimum}, GHE~\cite{abutaleb1989automatic}, and BPHEME~\cite{wang2005brightness}. There are also some deep-learning-based LLIE algorithms, such as MBLLEN~\cite{lv2018mbllen}, SGZSL~\cite{zheng2022semantic}, and DCC-Net~\cite{zhang2022deep}. However, applying LLIE algorithms directly to videos can lead to severe temporal instability. In order to fill the niche existing in LLIE, some LLVE algorithms that take temporal consistency into account are proposed, such as MBLLVEN~\cite{lv2018mbllen}, SDSD~\cite{wang2021seeing}, SMID~\cite{chen2019seeing}, and StableLLVE~\cite{zhang2021learning}.

Video Quality Assessment (VQA) is of great significance to facilitate the development of LLVE algorithms. 
Objective VQA 
can be divided into Full-Reference (FR) VQA~\cite{bampis2018spatiotemporal}, Reduced-Reference (RR) VQA~\cite{soundararajan2012video}, and No-Reference (NR) VQA~\cite{saad2014blind} contingent on the amount of
required pristine video information~\cite{sun2022deep}. Due to the difficulty in obtaining reference videos, NR-VQA has attracted a large number of researchers' attention. In the early development stages of NR-VQA, researchers often evaluate video quality based on handcrafted features~\cite{tu2020comparative, saad2014blind, mittal2015completely, korhonen2019two, tu2021ugc, korhonen2020blind, tu2021rapique}, such as structure, texture, and statistical features. Recently, owing to the potential in practical applications, deep learning based NR-VQA models~\cite{li2022blindly, li2019quality, wang2021rich, xu2021perceptual, ying2021patch, sun2022deep, wu2022fast} have progressively dominated the VQA field. However, most existing VQA models are designed for general purpose. To the best of our knowledge, few models specifically evaluate the quality of videos enhanced by LLVE algorithms. One possible reason is the lack of corresponding datasets. 

\begin{figure*}[ht]
\setlength{\abovecaptionskip}{0cm} 
\setlength{\belowcaptionskip}{0.1cm}
\centering
\includegraphics[width=\textwidth]{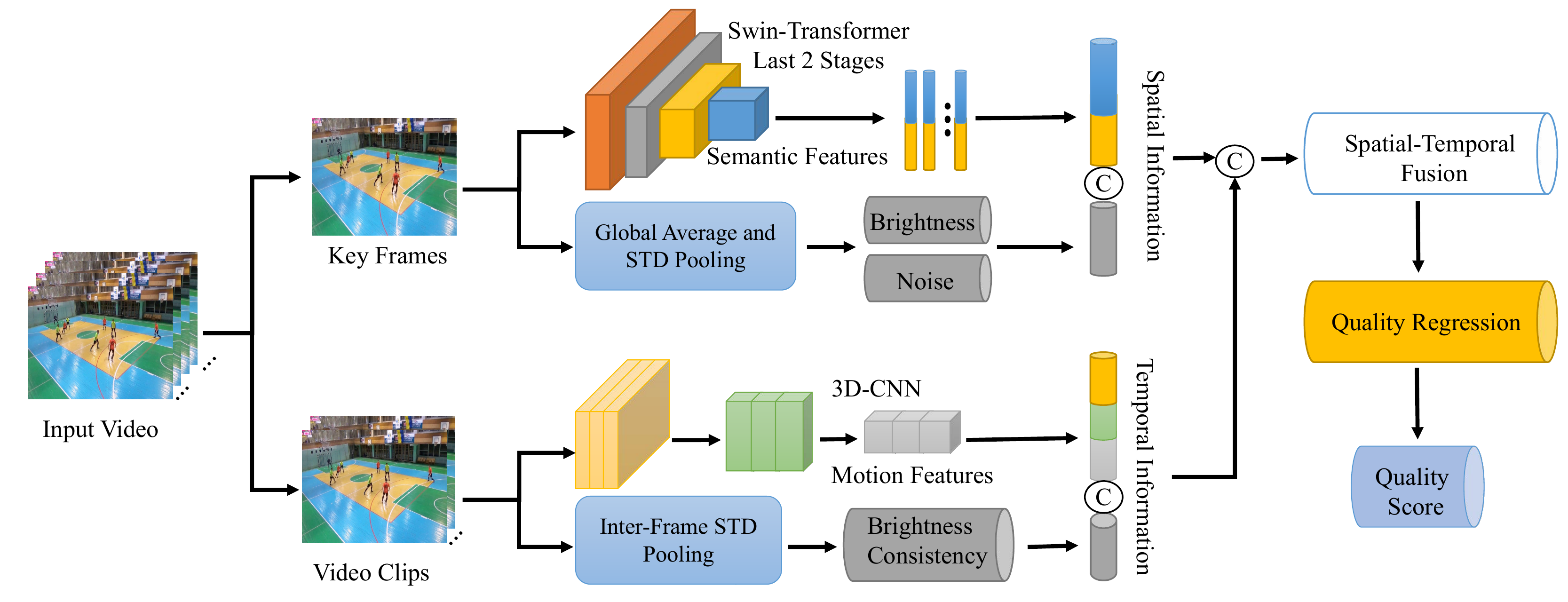}
\caption{Framework of Light-VQA. The model contains the spatial and temporal information extraction module, the feature fusion module, and the quality regression module. Concretely, spatial information contains semantic features, brightness, and noise. Temporal information contains motion features and brightness consistency.}
\label{architecture}
\vspace{-1em}
\end{figure*}

Therefore, in this paper, we elaborately build a Low-Light Video Enhancement
Quality Assessment (LLVE-QA) dataset to facilitate the work on evaluating the performance of LLVE algorithms. Different from general datasets which commonly consist of original UGC videos with various degradations,
LLVE-QA dataset contains 254 original low-light videos and 1,806 enhanced videos from representative enhancement algorithms, each with a corresponding MOS. Subsequently, we propose a quality assessment model specialized for low-light video enhancement, named Light-VQA. The framework of Light-VQA is illustrated in Figure \ref{architecture}. Considering that among low-level features, brightness and noise have the most impact on low-light enhanced VQA~\cite{zhai2021perceptual}, in addition to semantic features and motion features extracted from deep neural network, we specially handcraft the brightness, brightness consistency, and noise features to improve the ability of the model to represent the quality-aware features of low-light enhanced videos. Extensive experiments validate the effectiveness of our network design.

The contributions of this paper are summarized as follows:

\begin{enumerate}
\item By leveraging representative LLVE algorithms on the collected videos with diverse content and various degrees of brightness, we conduct a subjective experiment to build a low-light video enhancement dataset, named LLVE-QA. 
\item Benefiting from the built dataset, we propose a novel quality assessment model named Light-VQA \textit{specifically designed} for low-light enhanced videos that integrates the luminance-sensitive handcrafted features into deep-learning-based features in both spatial and temporal information, which is then fused to obtain the quality-aware representation. 
\item The proposed Light-VQA achieves the best performance as compared to 6 SOTA models on LLVE-QA and public dataset. We envision that the Light-VQA is promising to be a fundamental tool to assess the LLVE algorithms.
\end{enumerate}


\section{Related Work}
\subsection{Low-Light Enhancement}
To enhance the low-light videos, it is straightforward to split the low-light video into frames, so as to take advantage of existing LLIE algorithms. AGCCPF~\cite{gupta2016minimum} enhances the brightness and contrast of low-light images using the gamma correction and weighted probability distribution of pixels. GHE~\cite{abutaleb1989automatic} applies a transformation on image histogram to redistribute the pixel intensity, resulting in a more favorable visual result.
BPHEME~\cite{wang2005brightness} enhances the low-light video by balancing the brightness preserving histogram with maximum entropy. In addition to the above traditional algorithms, low-light image enhancement algorithms based on deep learning~\cite{lv2018mbllen, zheng2022semantic, zhang2022deep} are rapidly emerging. 
Zhang~\textit{et al.}~\cite{zhang2022deep} propose a consistent network to improve illumination and preserve color consistency of low-light images. However, applying LLIE algorithms directly to videos can lead to temporal consistency problems such as motion artifacts and brightness consistency, which will ultimately reduce the quality of videos.

In order to maintain the temporal consistency of videos better, specific LLVE algorithms~\cite{lv2018mbllen, wang2021seeing, chen2019seeing, zhang2021learning} are proposed. MBLLVEN~\cite{lv2018mbllen} processes low-light videos via 3D convolution to extract temporal information and preserve temporal consistency. Wang~\textit{et al.}~\cite{wang2021seeing} collect a new dataset that contains high-quality spatially-aligned video pairs in both low-light and normal-light conditions, and further design a self-supervised network to reduce noises and enhance the illumination based on the Retinex theory. Chen~\textit{et al.}~\cite{chen2019seeing} propose a siamese network and introduce a self-consistency loss to preserve color while suppressing spatial and temporal artifacts efficiently. StableLLVE~\cite{zhang2021learning} maintains the temporal consistency after enhancement by learning and inferring motion field (\textit{i.e.},optical flow) from synthesized short-range video sequences. In order to ensure the diversity of visual effects of the enhanced videos, we apply both LLIE and LLVE algorithms when constructing the LLVE-QA dataset.

\subsection{VQA Datasets}
In order to facilitate the development of VQA algorithms, many VQA datasets~\cite{ghadiyaram2017capture, sinno2018large, hosu2017konstanz, wang2019youtube, ying2021patch, li2020ugc, yu2021predicting, gao2023vdpve} have been proposed.
Videos in LIVE-Qualcomm~\cite{ghadiyaram2017capture} contain the following 6 distortion types: color, exposure, focus, artifacts, sharpness, and stabilization. LIVE-VQC~\cite{sinno2018large} contains 585 videos, which are captured by various cameras with different resolutions. In addition to the common distortions, the visual quality of UGC videos is influenced by compression generated while they are uploaded to and downloaded from the Internet. UGC-VIDEO~\cite{li2020ugc} and LIVE-WC~\cite{yu2021predicting} simulate the specific distortion by utilizing several video compression algorithms. 
KoNViD-1k~\cite{hosu2017konstanz}, YouTube-UGC~\cite{wang2019youtube}, and LSVQ~\cite{ying2021patch} extensively collect in-the-wild UGC videos from the Internet, greatly expand the scale of VQA datasets. Besides, VDPVE~\cite{gao2023vdpve} is constructed to fill in the gaps of VQA datasets specially for video enhancement, which can further promote the refined development of VQA models. However, most of existing datasets only contain unprocessed UGC videos containing various distortions. While VDPVE takes videos after enhancement into account, it is still general and not targeted. Our LLVE-QA dataset focuses on original low-light videos and corresponding enhanced videos after LLVE, which lays a solid foundation for designing a specific LLVE quality assessment model.  
    

\subsection{NR-VQA Models}

\begin{figure*}[ht]
\setlength{\abovecaptionskip}{0cm} 
\setlength{\belowcaptionskip}{0.1cm}
\centering
\begin{subfigure}[t]{0.15\textwidth}
    \includegraphics[width=1\textwidth]{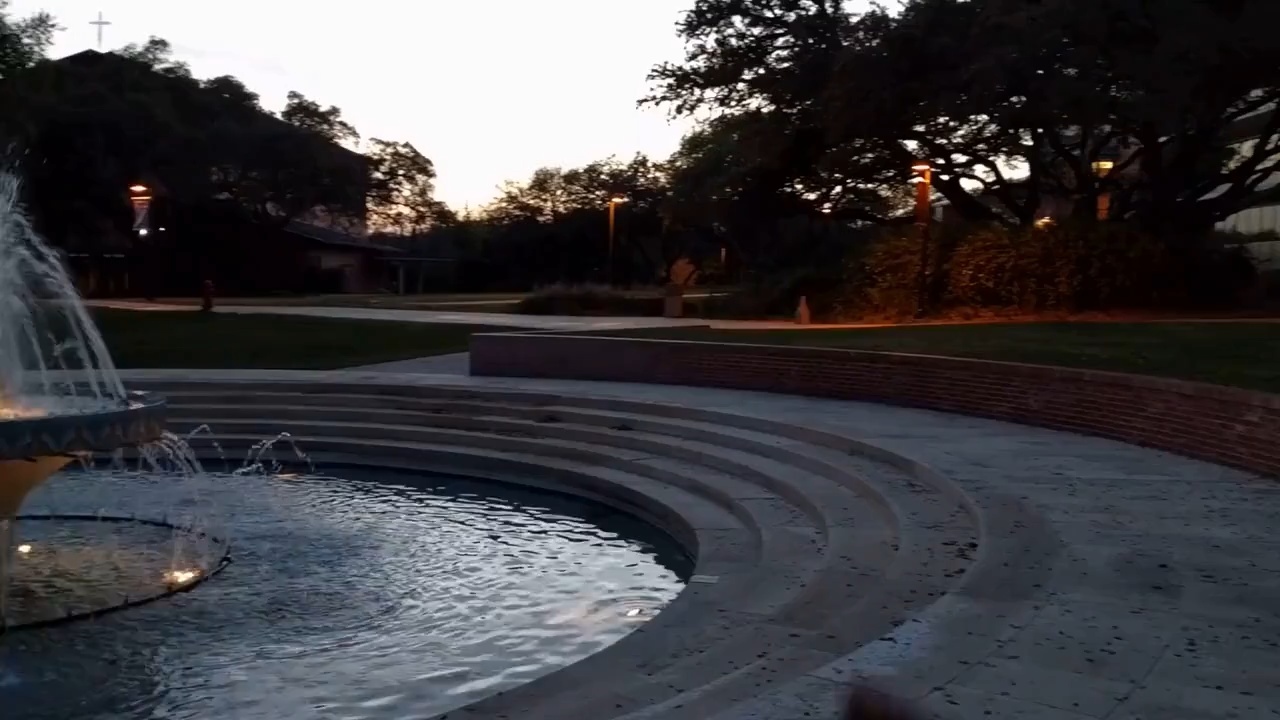}
    \subcaption*{(a)\ Original}
\end{subfigure}
\begin{subfigure}[t]{0.15\textwidth}
    \includegraphics[width=1\textwidth]{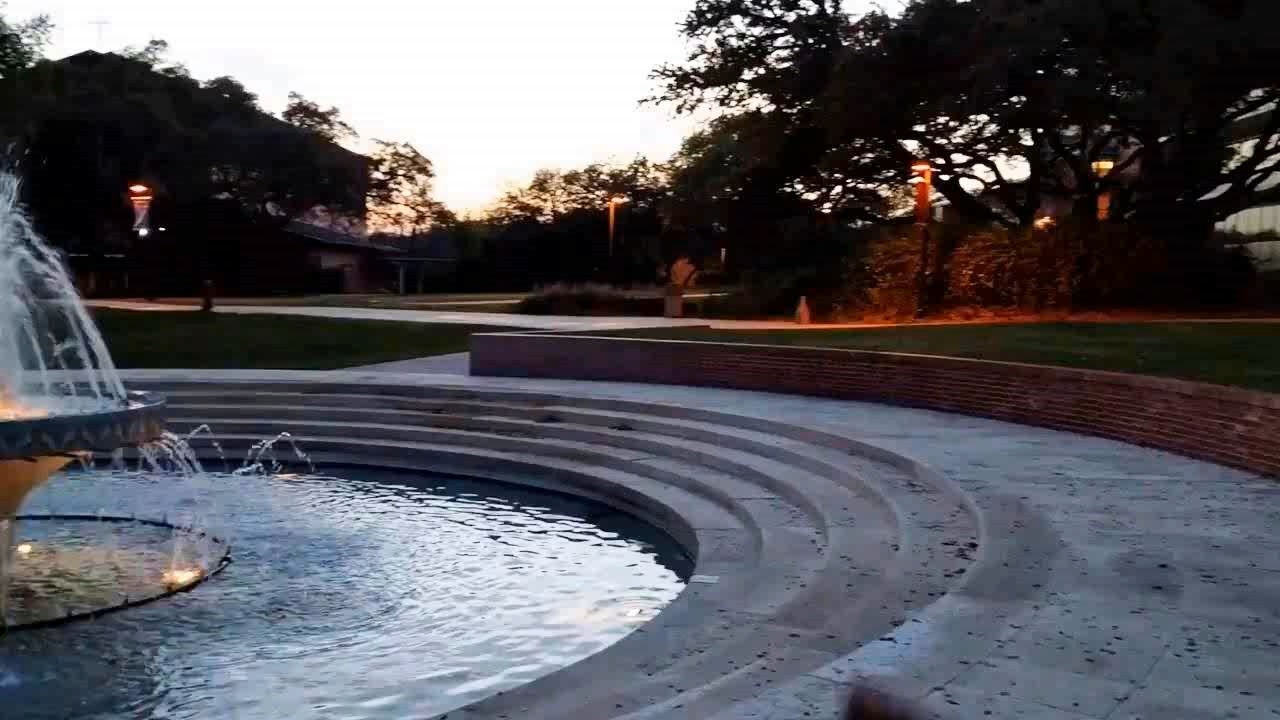}
    \subcaption*{(b)\ AGCCPF}
\end{subfigure}
\begin{subfigure}[t]{0.15\textwidth}
    \includegraphics[width=1\textwidth]{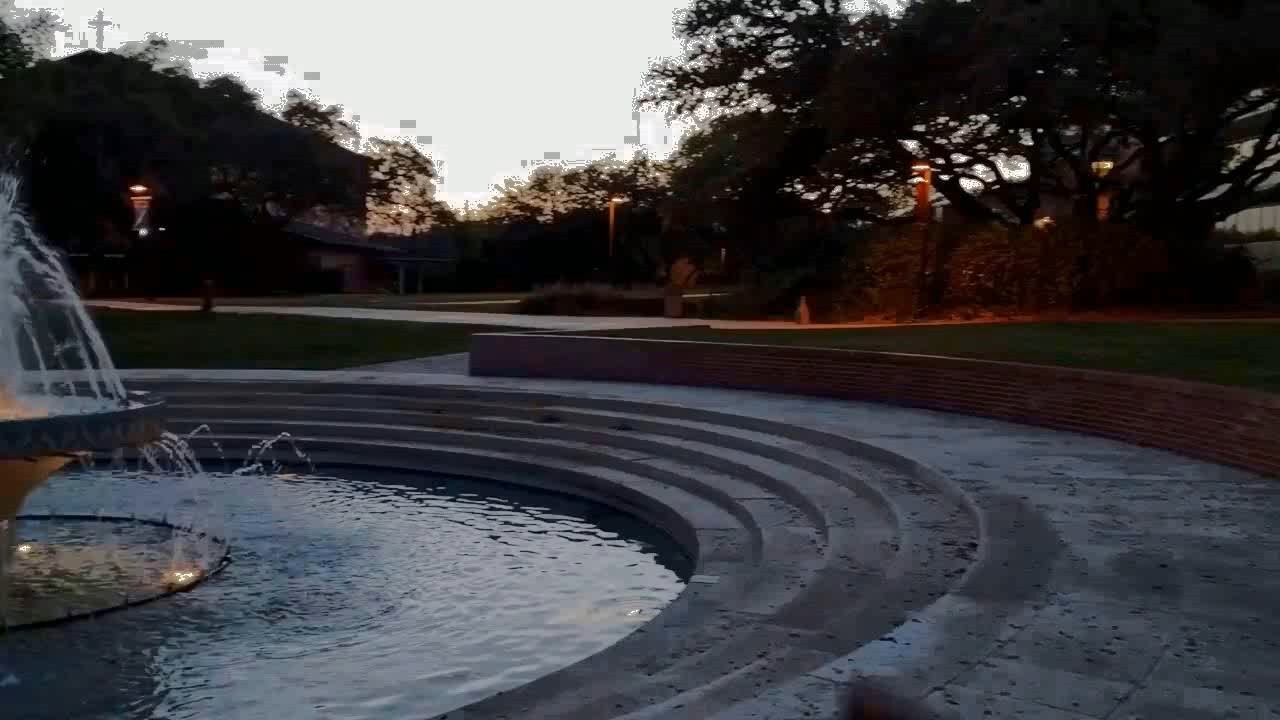}
    \subcaption*{(c)\ BPHEME}
\end{subfigure}
\quad
\begin{subfigure}[t]{0.15\textwidth}
    \includegraphics[width=1\textwidth]{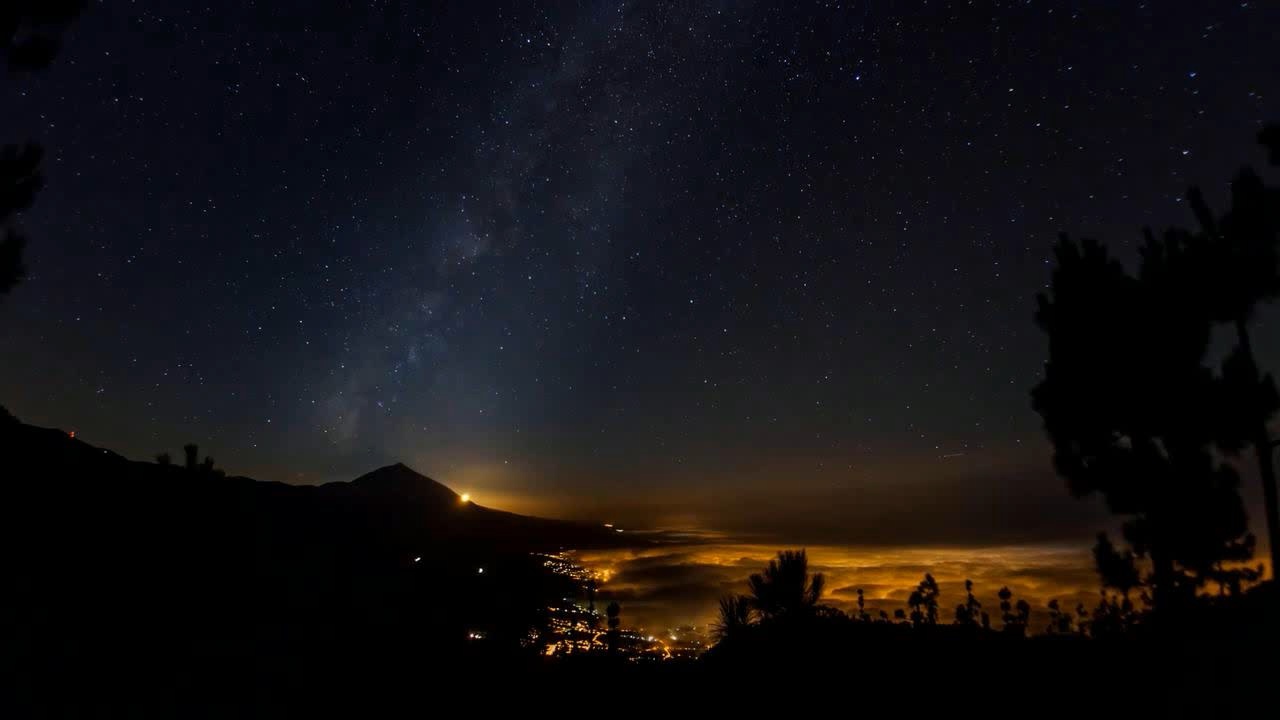}
    \subcaption*{(a)\ Original}
\end{subfigure}
\begin{subfigure}[t]{0.15\textwidth}
    \includegraphics[width=1\textwidth]{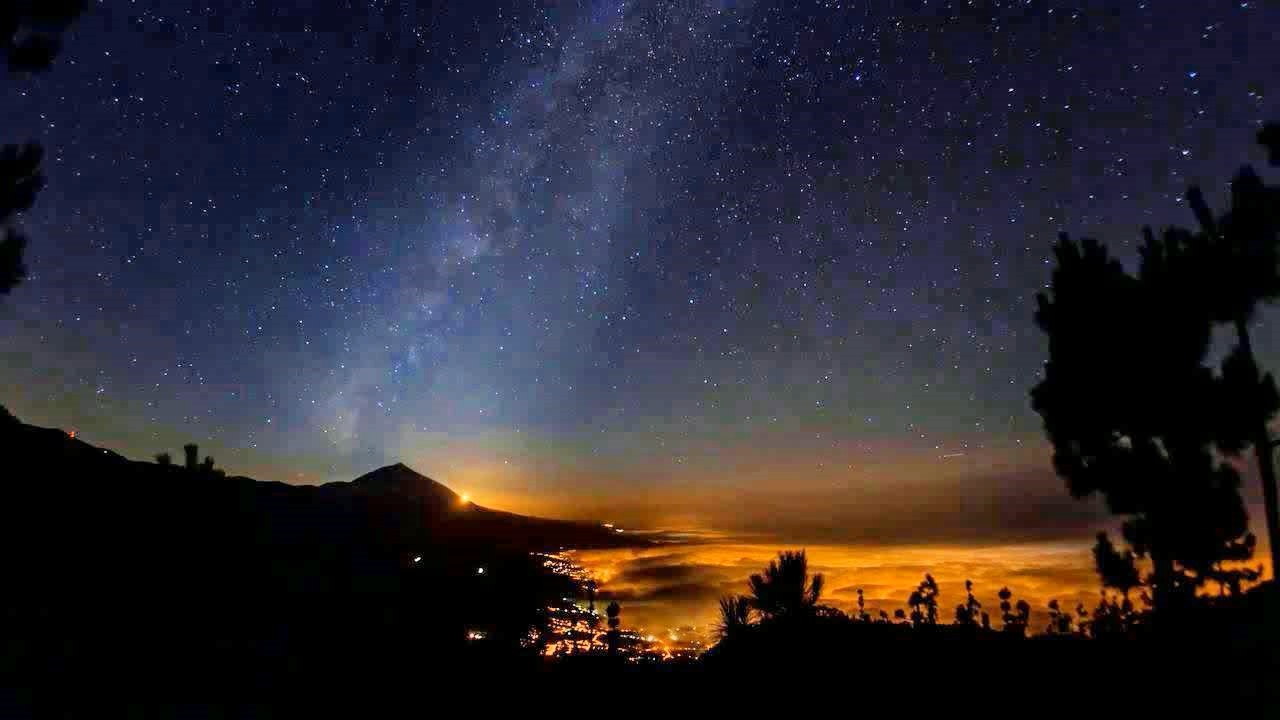}
    \subcaption*{(b)\ AGCCPF}
\end{subfigure}
\begin{subfigure}[t]{0.15\textwidth}
    \includegraphics[width=1\textwidth]{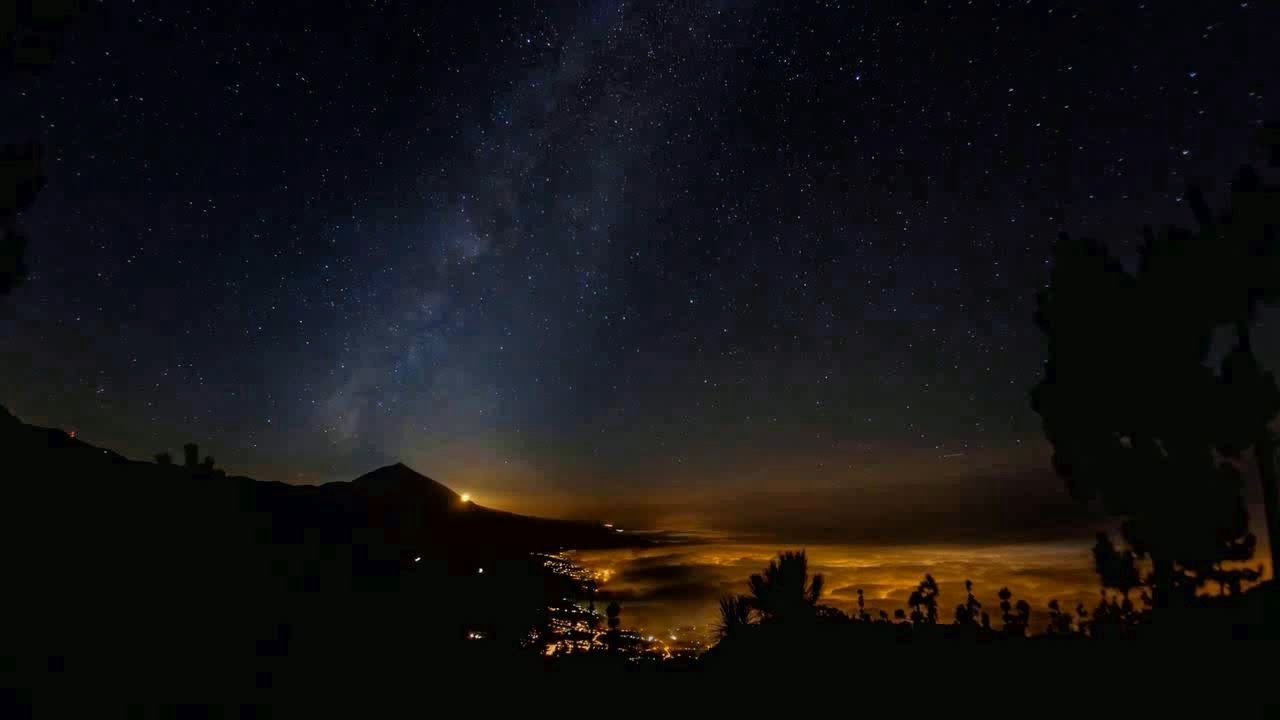}
    \subcaption*{(c)\ BPHEME}
\end{subfigure}
\begin{subfigure}[t]{0.15\textwidth}
    \includegraphics[width=1\textwidth]{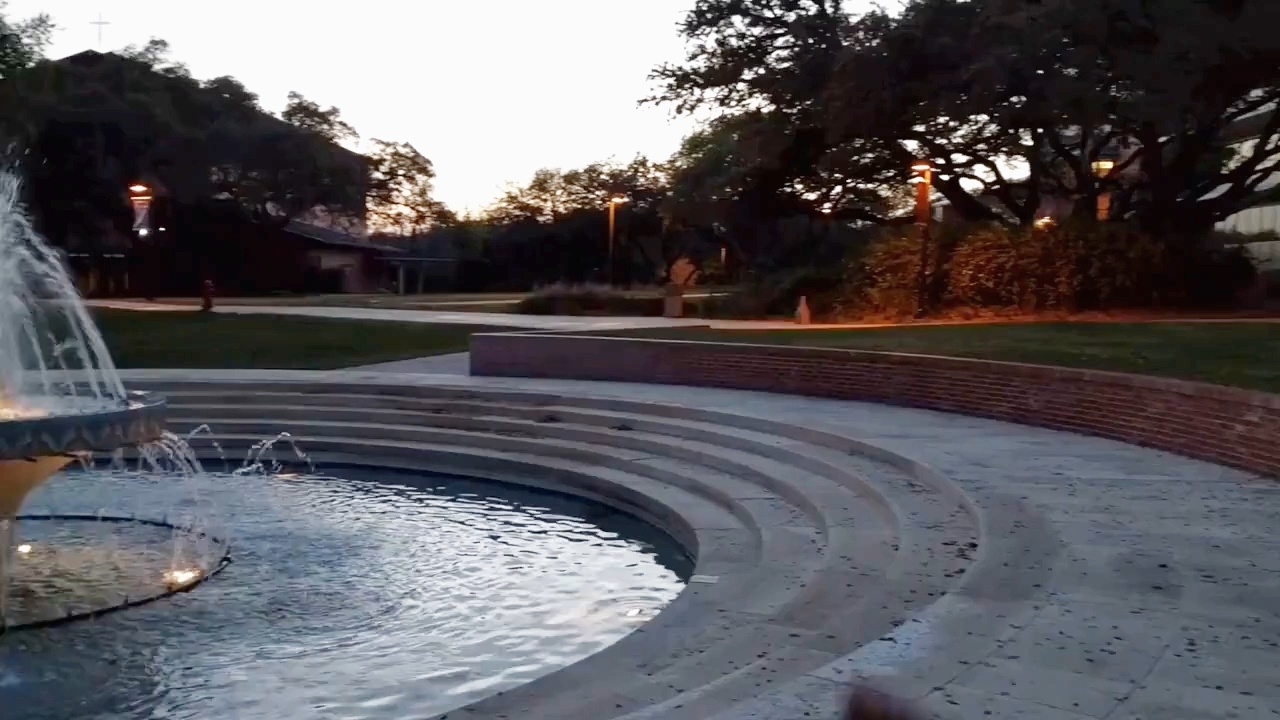}
    \subcaption*{(d)\ CapCut}
\end{subfigure}
\begin{subfigure}[t]{0.15\textwidth}
    \includegraphics[width=1\textwidth]{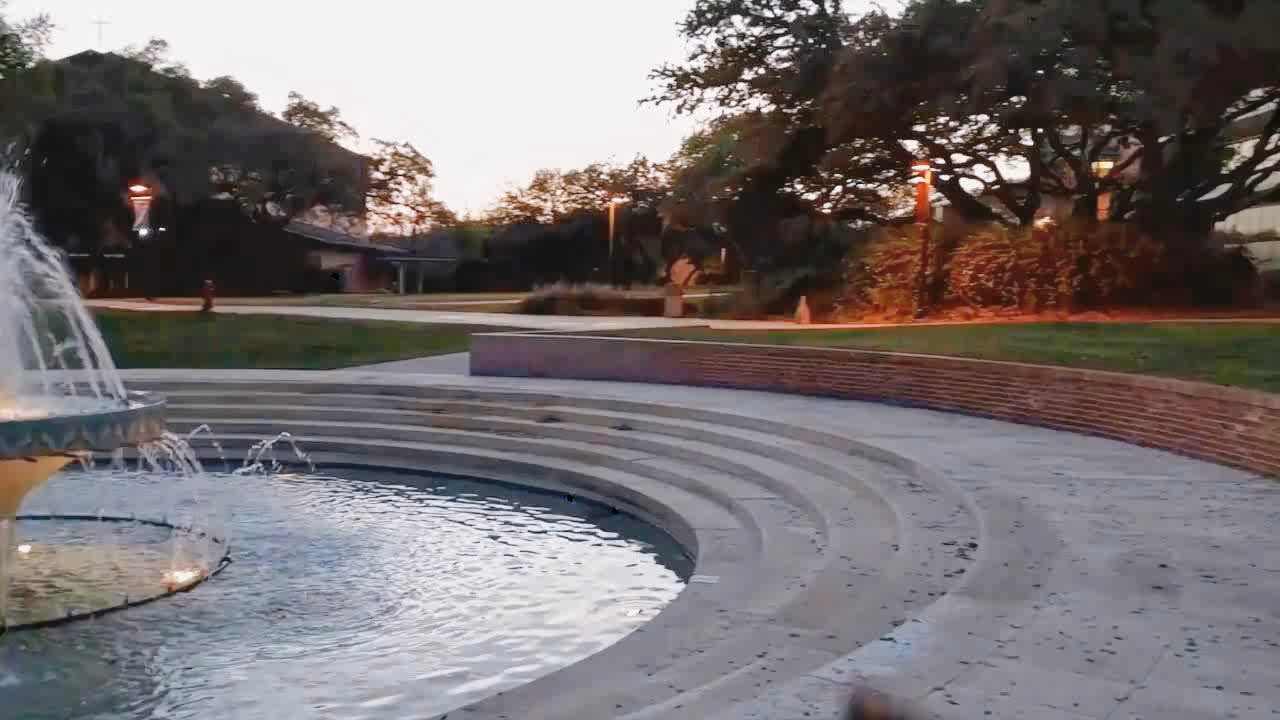}
    \subcaption*{(e)\ DCC-Net}
\end{subfigure}
\begin{subfigure}[t]{0.15\textwidth}
    \includegraphics[width=1\textwidth]{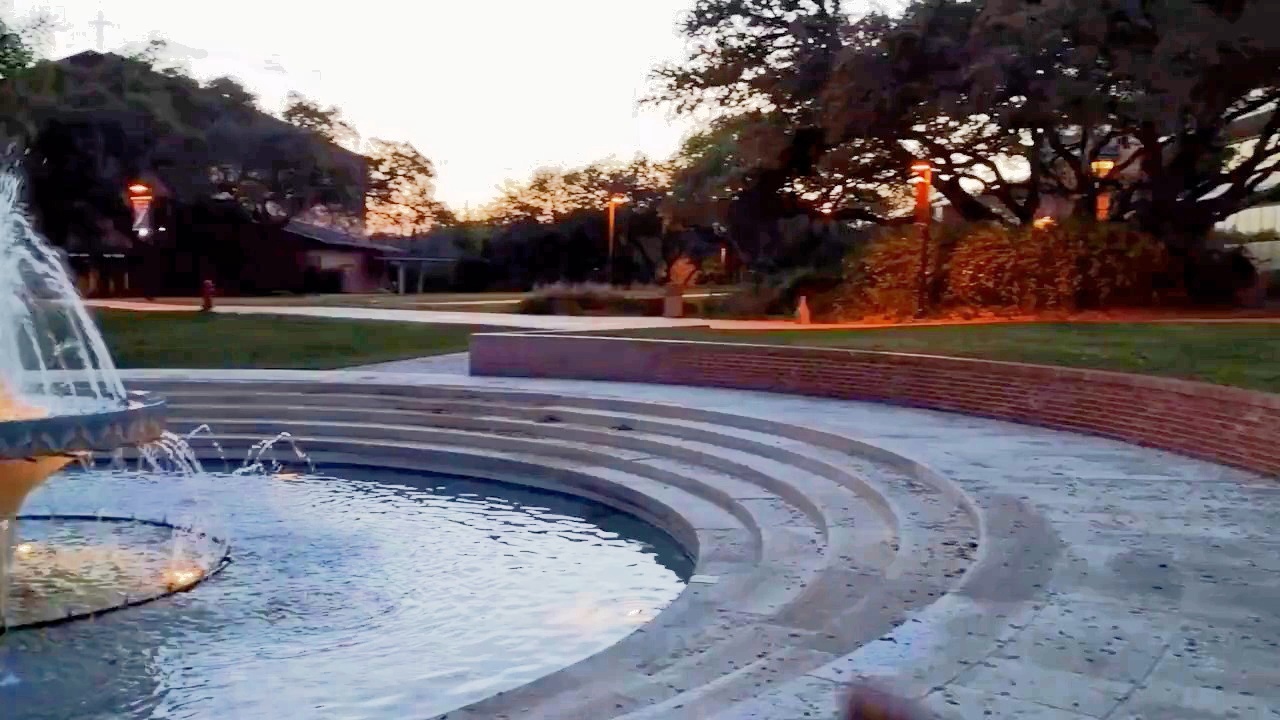}
    \subcaption*{(f)\ GHE}
\end{subfigure}
\quad
\begin{subfigure}[t]{0.15\textwidth}
    \includegraphics[width=1\textwidth]{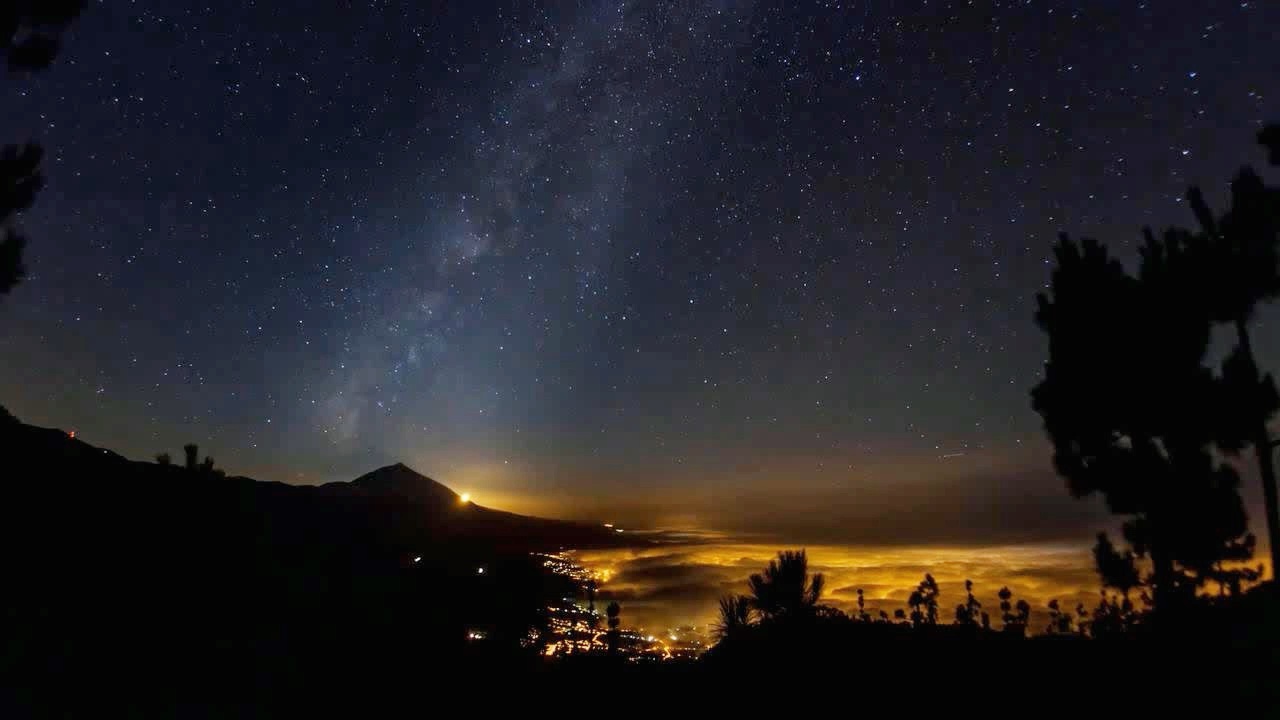}
    \subcaption*{(d)\ CapCut}
\end{subfigure}
\begin{subfigure}[t]{0.15\textwidth}
    \includegraphics[width=1\textwidth]{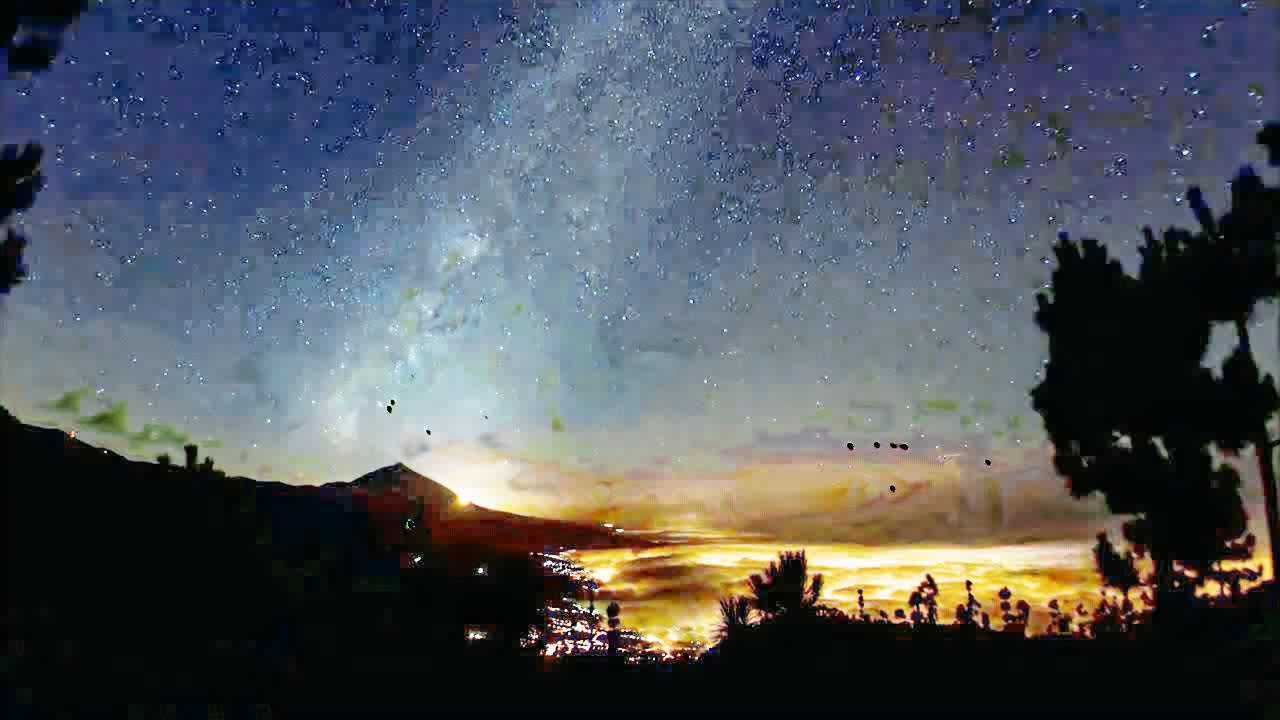}
    \subcaption*{(e)\ DCC-Net}
\end{subfigure}
\begin{subfigure}[t]{0.15\textwidth}
    \includegraphics[width=1\textwidth]{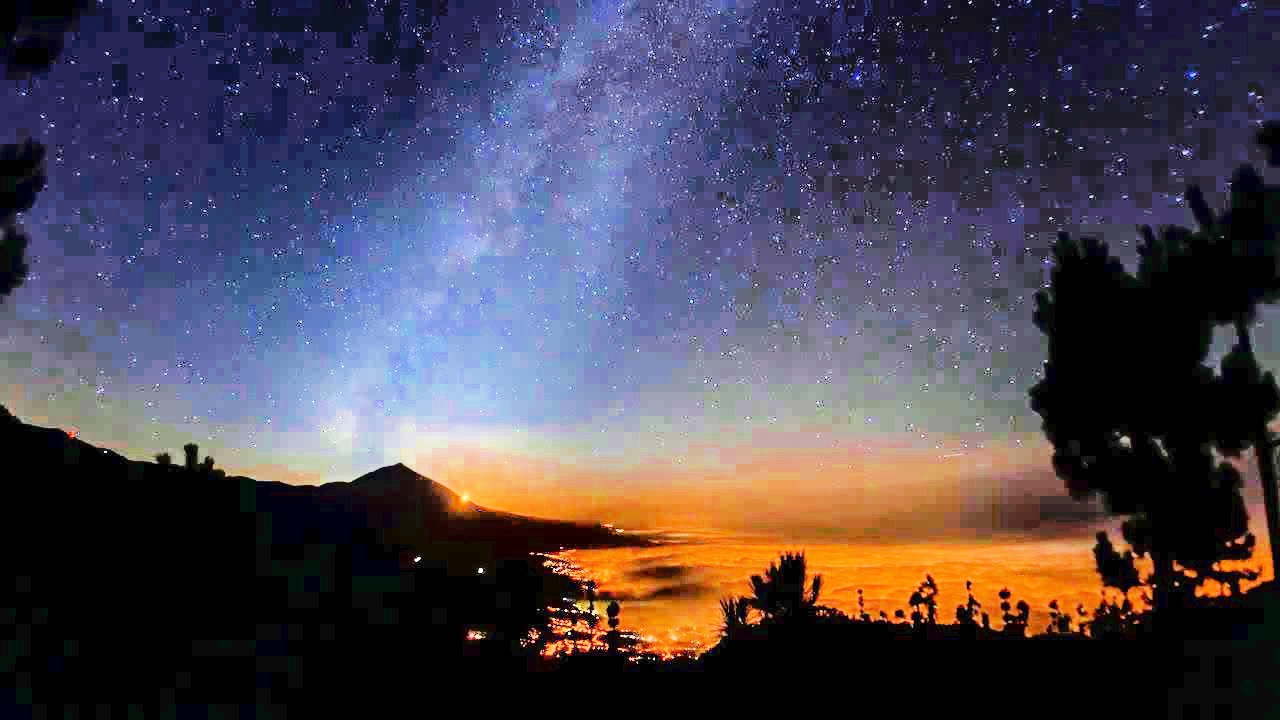}
    \subcaption*{(f)\ GHE}
\end{subfigure}
\begin{subfigure}[t]{0.15\textwidth}
    \includegraphics[width=1\textwidth]{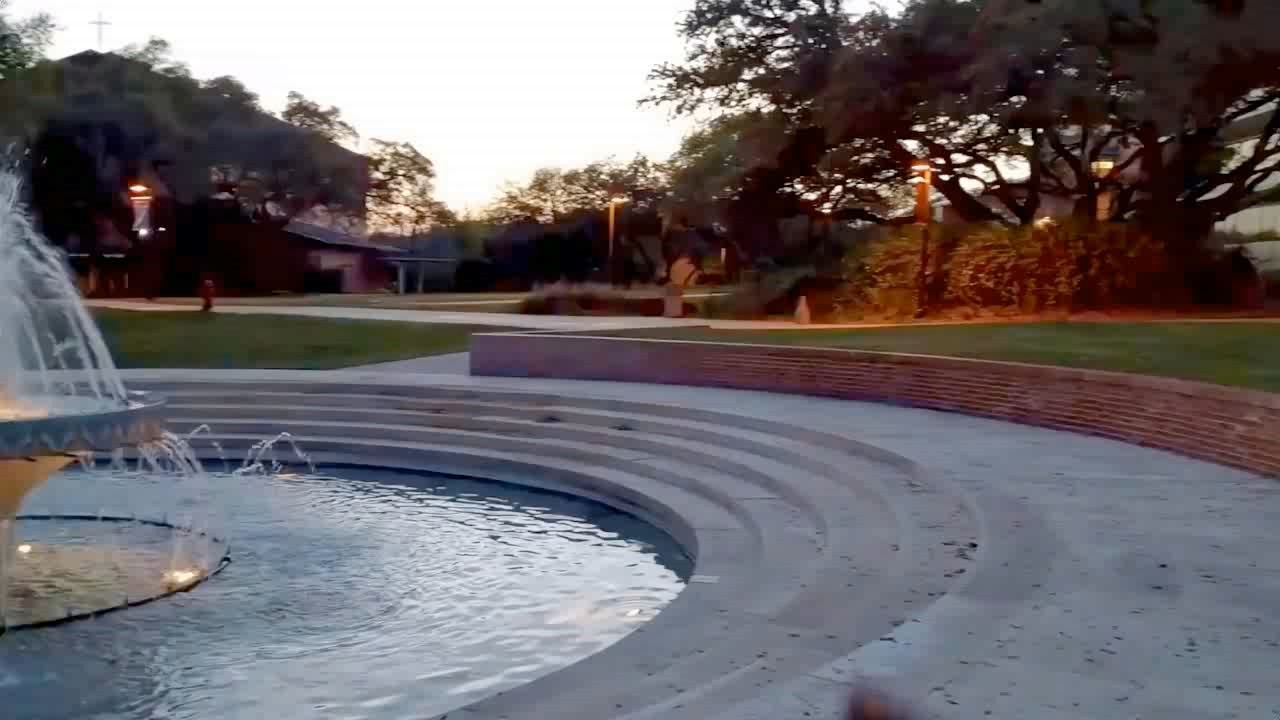}
    \subcaption*{(g)\ MBLLVEN}
\end{subfigure}
\begin{subfigure}[t]{0.15\textwidth}
    \includegraphics[width=1\textwidth]{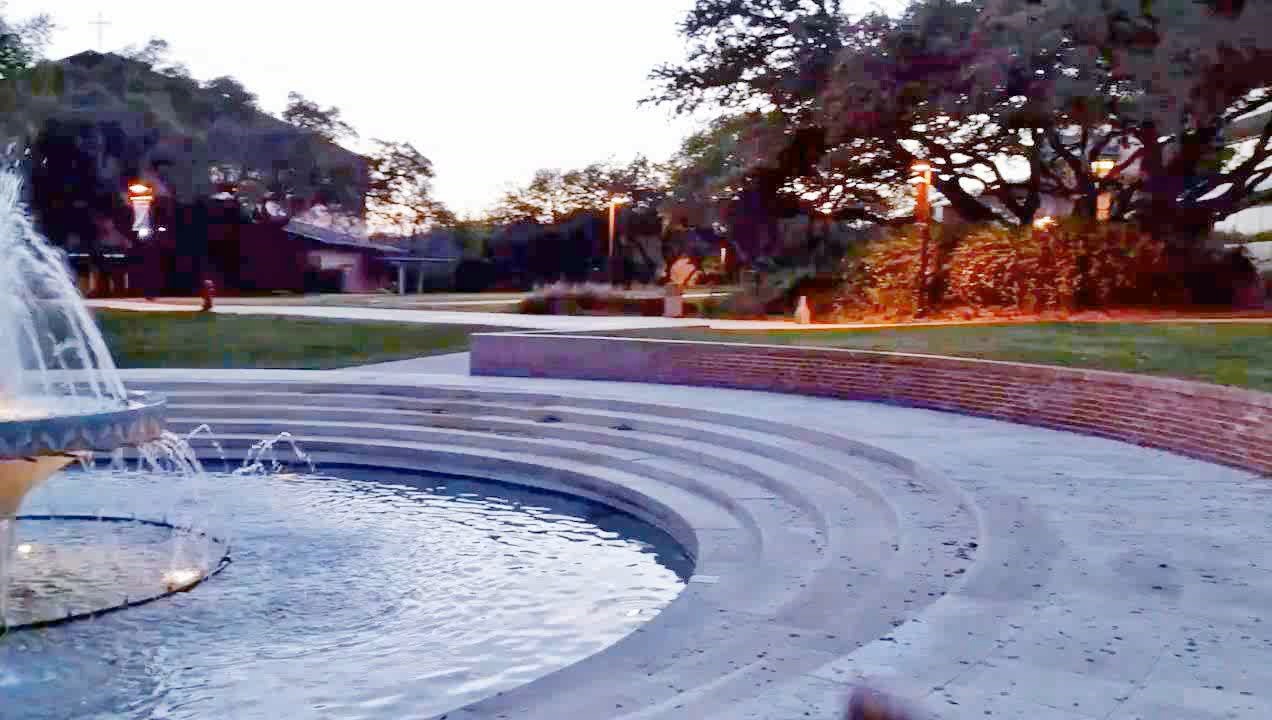}
    \subcaption*{(h)\ SGZSL}
\end{subfigure}
\begin{subfigure}[t]{0.15\textwidth}
    \includegraphics[width=1\textwidth]{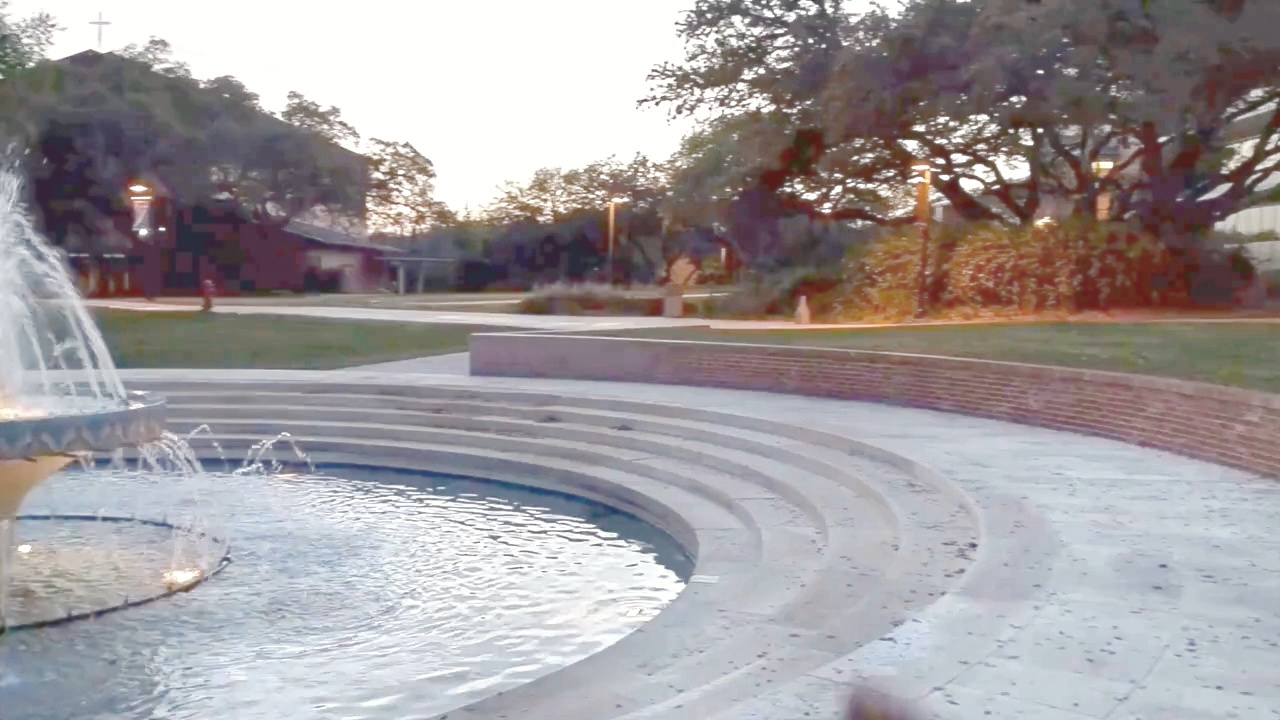}
    \subcaption*{(i)\ StableLLVE}
\end{subfigure}
\quad
\begin{subfigure}[t]{0.15\textwidth}
    \includegraphics[width=1\textwidth]{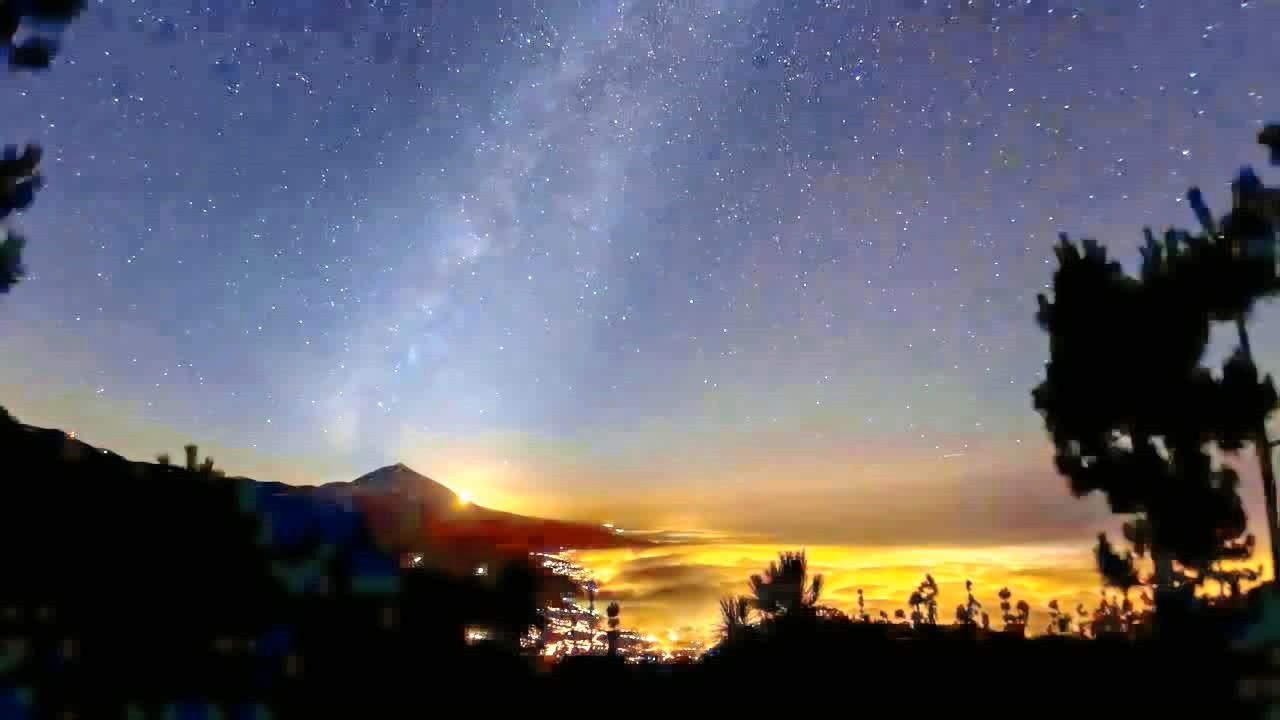}
    \subcaption*{(g)\ MBLLVEN}
\end{subfigure}
\begin{subfigure}[t]{0.15\textwidth}
    \includegraphics[width=1\textwidth]{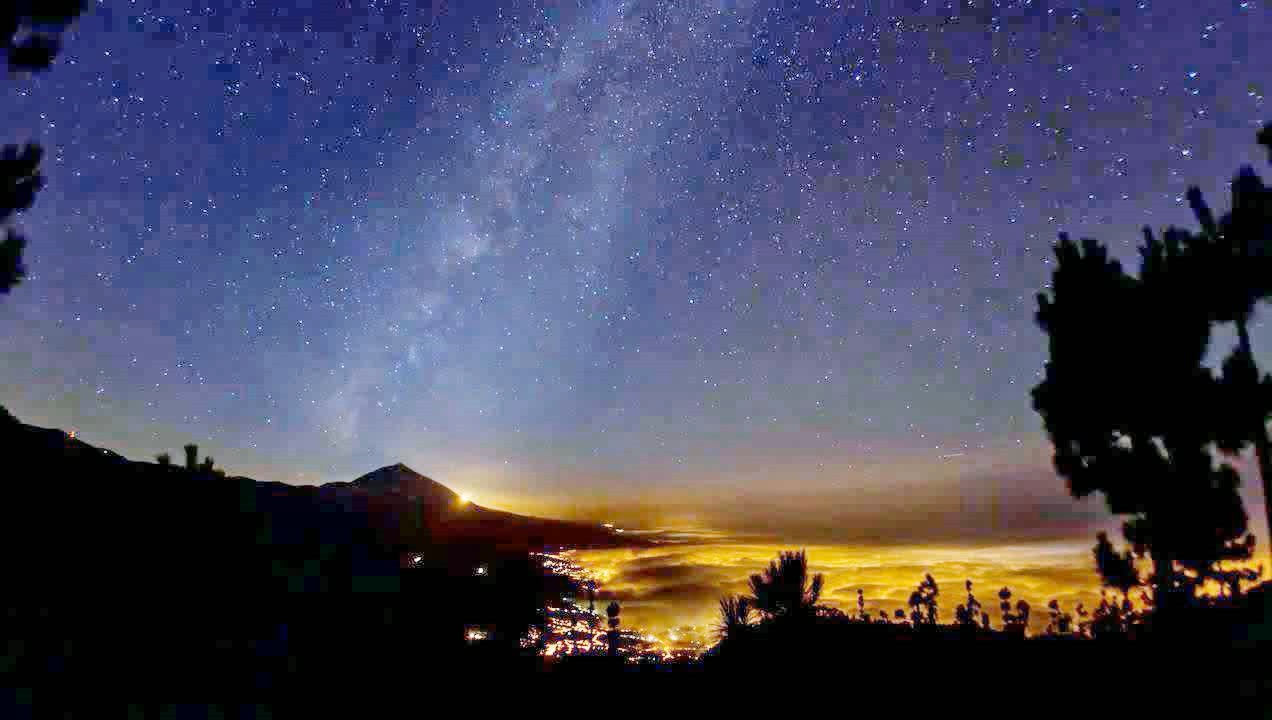}
    \subcaption*{(h)\ SGZSL}
\end{subfigure}
\begin{subfigure}[t]{0.15\textwidth}
    \includegraphics[width=1\textwidth]{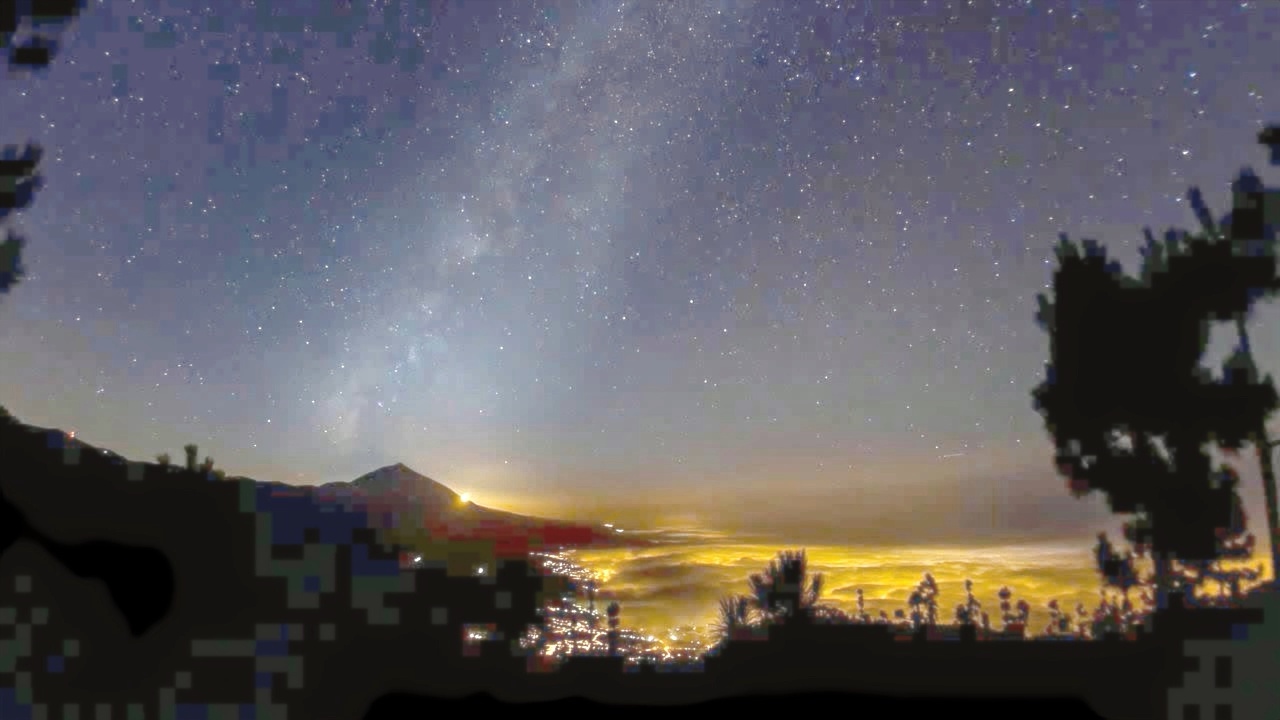}
    \subcaption*{(i)\ StableLLVE}
\end{subfigure}
\caption{Representative frames of two original videos and their corresponding enhanced videos.}
\label{dataset}
\vspace{-1em}
\end{figure*}

The traditional and naive NR-VQA models are based on handcrafted features~\cite{saad2014blind, mittal2015completely, korhonen2019two, tu2021ugc, korhonen2020blind}. These handcrafted features, including spatial features, temporal features, statistical features, and so on, can be extracted to learn the quality scores of videos. For example, V-BLIINDS~\cite{saad2014blind} builds a Natural Scene Statistics (NSS) module to extract spatial-temporal features and a motion module to quantify motion coherency. The core of TLVQM~\cite{korhonen2019two} is to generate video features in two levels, in which low complexity features are extracted from the full sequence first, and then high complexity features are extracted in key frames which are selected by utilizing low complexity features.  VIDEVAL~\cite{tu2021ugc} combines existing VQA methods together and proposes a feature selection strategy, which can choose appropriate features and then fuse them efficiently to predict the quality scores of videos.

With the rapid pace of technological advancements, VQA models based on deep learning~\cite{li2022blindly, li2019quality, wang2021rich, yi2021attention, ying2021patch, sun2022deep, wu2022fast, tu2021rapique, stablevqa, digitalqa, wu2022fastervqa, wu2023dover, wu2023explainablevqa, wu2023bvqiplus, wu2023bvqi, wu2023discovqa} have progressively emerged as the prevailing trend. For example, based on a pre-trained DNN model and Gated Recurrent Units (GRUs), VSFA~\cite{li2019quality} reflects the temporal connection between the semantic features of key frames well. BVQA~\cite{li2022blindly} and Simple-VQA~\cite{sun2022deep} further take the impact of motion features on videos into account and introduce motion features extracted by the pre-trained 3D CNN models. Wang~\textit{et al.}~\cite{wang2021rich} propose a DNN-based framework to measure the quality of UGC videos from three aspects: video content, technical quality, and compression level. FAST-VQA~\cite{wu2022fast} creatively introduces a Grid Mini-patch Sampling to generate fragments, and utilizes a model with Swin Transformer~\cite{liu2021swin} as the backbone to extract features efficiently from these fragments. RAPIQUE~\cite{tu2021rapique} leverages quality-aware statistical features and semantics-aware convolutional features, which first attempts to combine handcrafted and deep-learning-based features. While prior VQA models are designed for general UGC videos without exception, our Light-VQA model focuses on LLVE quality assessment by additionally introducing handcrafted brightness and noise features that significantly affect the quality of low-light videos and their corresponding enhanced results to improve the assessment accuracy. 

\section{Dataset Preparation}

\subsection{Video Collection}
A high-quality dataset is a prerequisite for a well-performing model. To start with,
we elaborately select 254 low-light videos from VDPVE~\cite{gao2023vdpve},  LIVE-VQC~\cite{sinno2018large}, YouTube-UGC~\cite{wang2019youtube}, and SDSD~\cite{wang2021seeing} datasets. The low-light videos we choose contain diverse content and various degrees of brightness. 
Subsequently, 
we employ 7 low light enhancement algorithms (\textit{i.e.}, AGCCPF~\cite{gupta2016minimum}, GHE~\cite{abutaleb1989automatic}, BPHEME~\cite{wang2005brightness}, SGZSL~\cite{zheng2022semantic}, DCC-Net~\cite{zhang2022deep}, MBLLVEN~\cite{lv2018mbllen} and StableLLVE~\cite{zhang2021learning}) and one commercial software CapCut~\cite{capcut-web} respectively to obtain the enhanced videos. We further remove the videos with extremely poor visual quality due to the distortions generated in the process of enhancement. Eventually, 254 original low-light videos and 1,806 enhanced videos constitute our LLVE-QA dataset. To the best of our knowledge, this is the \textit{first} dataset specifically designed for evaluating low-light video enhancement algorithms. Representative frames of two original videos and their corresponding enhanced videos are shown in Figure \ref{dataset}.

\subsection{Subjective Experiment}
We invite 22 subjects to participate in the subjective experiment. All of them are professional and experienced data labeling staff. Subjects are required to score the quality of videos within the range of [0, 100]. The scoring criteria are that higher score corresponds to the better video quality.
In the process of scoring a group of videos (including an original low-light video and corresponding enhanced videos), to make subjects not limited to the video content but pay more attention to the visual perceptual quality of the videos, we customize a scoring interface which is demonstrated in Supplementary. 
Subjects are supposed to score the original video first. When they score the enhanced videos, they can observe and compare them to the original video repeatedly. 
Compared to randomly shuffling the order of videos for scoring, the subjective quality scores obtained in this way can better reflect the visual perceptual difference caused by LLVE.

\subsection{Data Analysis}

\begin{figure}[t]
\setlength{\abovecaptionskip}{0cm} 
\setlength{\belowcaptionskip}{0.1cm}
  \centering
  \begin{subfigure}[b]{0.49\linewidth}
    \centering
    \includegraphics[width=\linewidth]{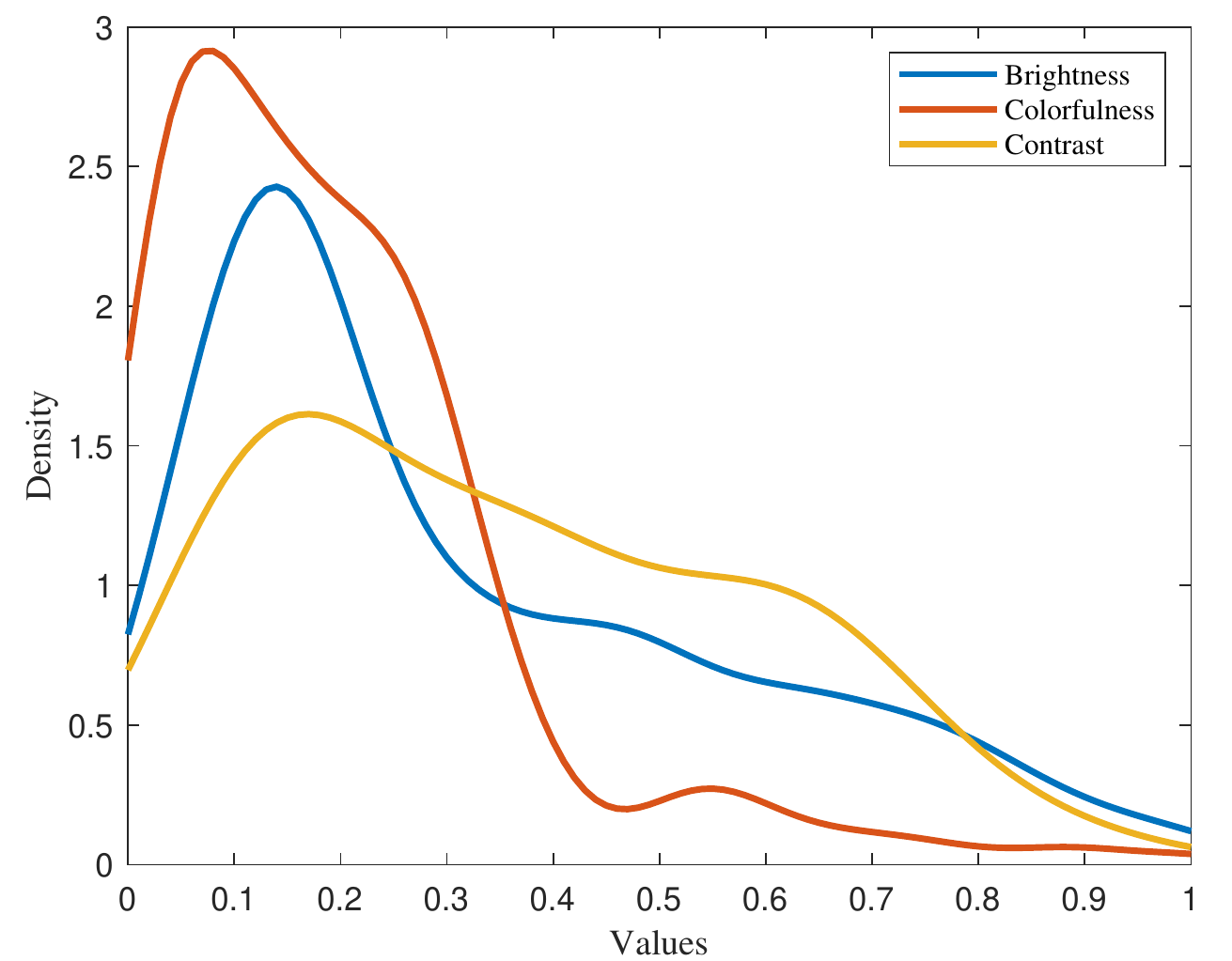}
    \caption{Original videos}
    \label{or_index}
  \end{subfigure}
  \hfill
  \begin{subfigure}[b]{0.49\linewidth}
    \centering
    \includegraphics[width=\linewidth]{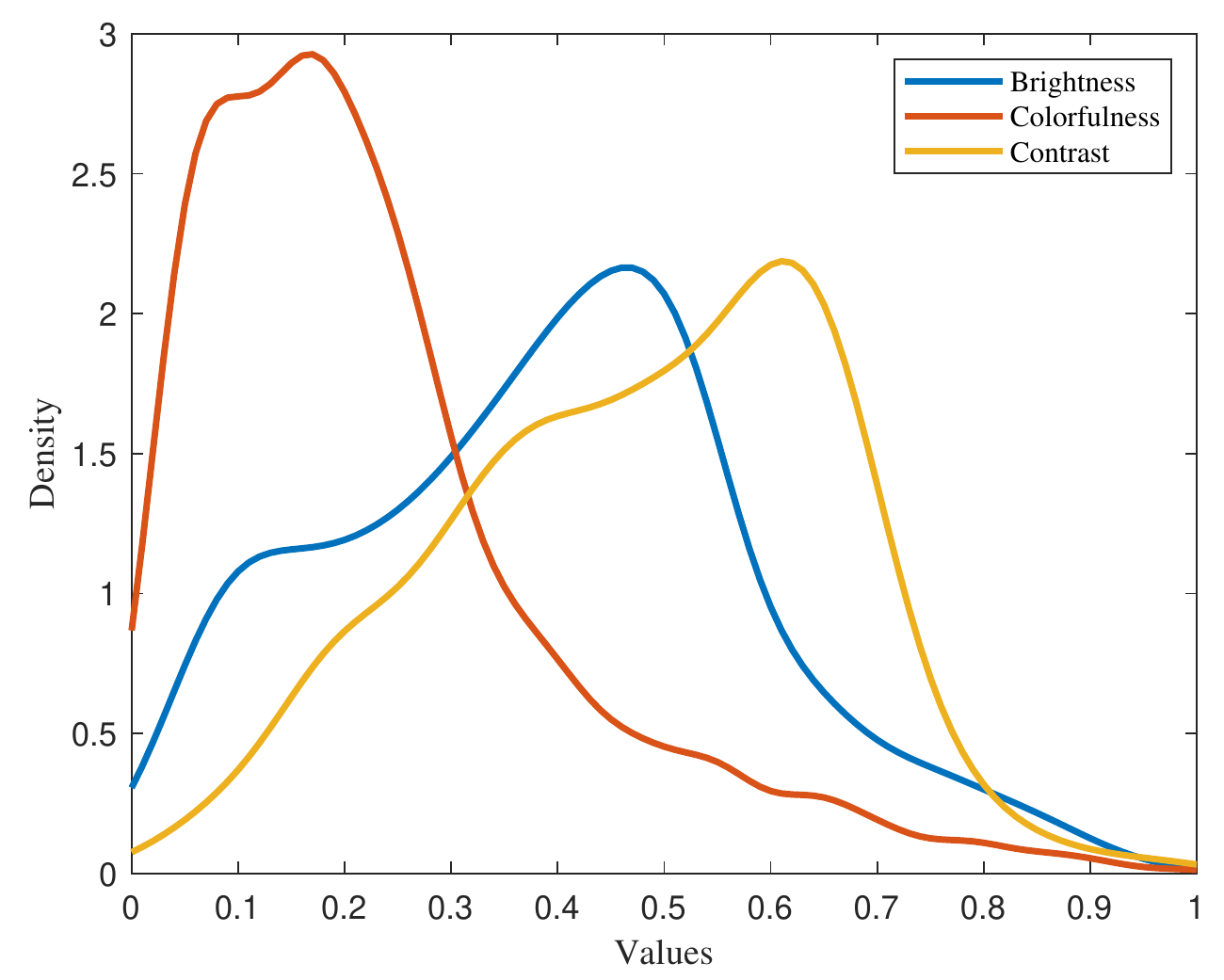}
    \caption{Enhanced videos}
    \label{en_index}
  \end{subfigure}
  \caption{Distributions of brightness, contrast, and colorfulness over the original and enhanced videos in our LLVE-QA dataset.}
  \label{index}
  \vspace{-1em}
\end{figure}

In order to measure the visual perceptual difference between original and enhanced videos, we calculate three video attributes~\cite{gao2023vdpve}: brightness, contrast, and colorfulness, which are normalized and shown in Figure \ref{index}. Colorfulness is not significantly changed before and after video enhancement, while the brightness and contrast have undergone major changes, which is in line with visual perception. Since there is a large amount of redundant information between adjacent frames, we only select a subset of all video frames for processing. The concrete calculation process are listed as follows~\cite{hosu2017konstanz}:

\textit{(1) Brightness: }Given a video frame, we convert it to grayscale and compute the average of pixel values. Then the brightness result of a video is obtained by averaging the brightness of all selected frames.

\textit{(2) Contrast: }For a video frame, its contrast is obtained simply by computing standard deviation of pixel grayscale intensities~\cite{peli1990contrast}. Then we average the contrast results of all selected frames to get the contrast of a video.

\textit{(3) Colorfulness: }We utilize Hasler and Suesstrunk’s metric~\cite{hasler2003measuring} to calculate this attribute. Specifically, given a video frame in RGB format, we compute $rg = R - G$ and $yb = \frac{1}{2}(R + G) - B$ first, and then the colorfulness is calculated by $\sqrt{\sigma_{rg}^{2} + \sigma_{yb}^{2}} + \frac{3}{10}\sqrt{\mu_{rg}^{2} + \mu_{yb}^{2}}$, where $\sigma^{2}$ and $\mu$ represent the variance and mean value respectively. Finally, we average the colorfulness values of all selected frames to obtain the colorfulness of a video.

\begin{figure}[t]
\setlength{\abovecaptionskip}{0cm} 
\setlength{\belowcaptionskip}{0.1cm}
  \centering
  \begin{subfigure}[b]{0.49\linewidth}
    \centering
    \includegraphics[width=\linewidth]{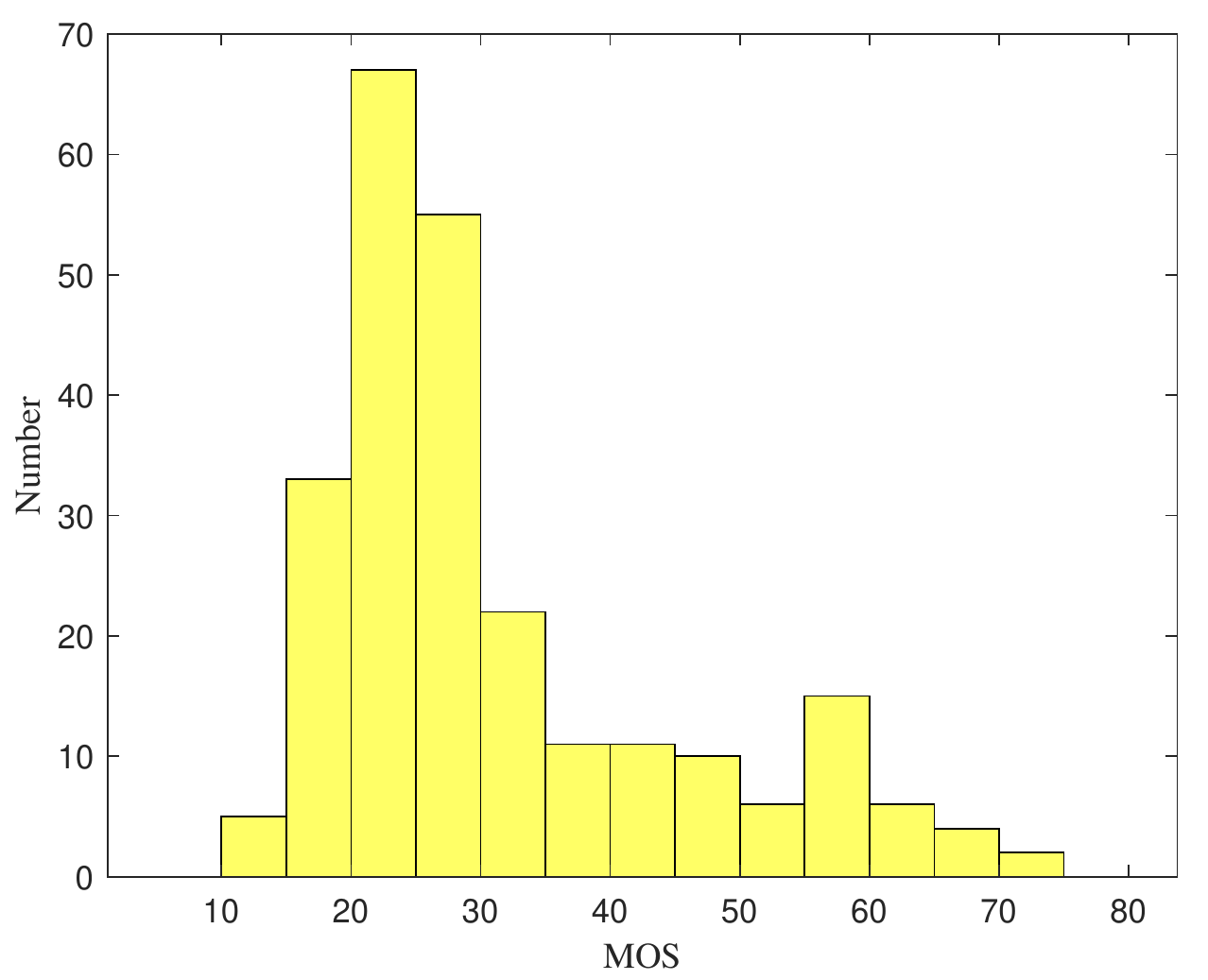}
    \caption{Original videos}
    \label{h1}
  \end{subfigure}
  \hfill
  \begin{subfigure}[b]{0.49\linewidth}
    \centering
    \includegraphics[width=\linewidth]{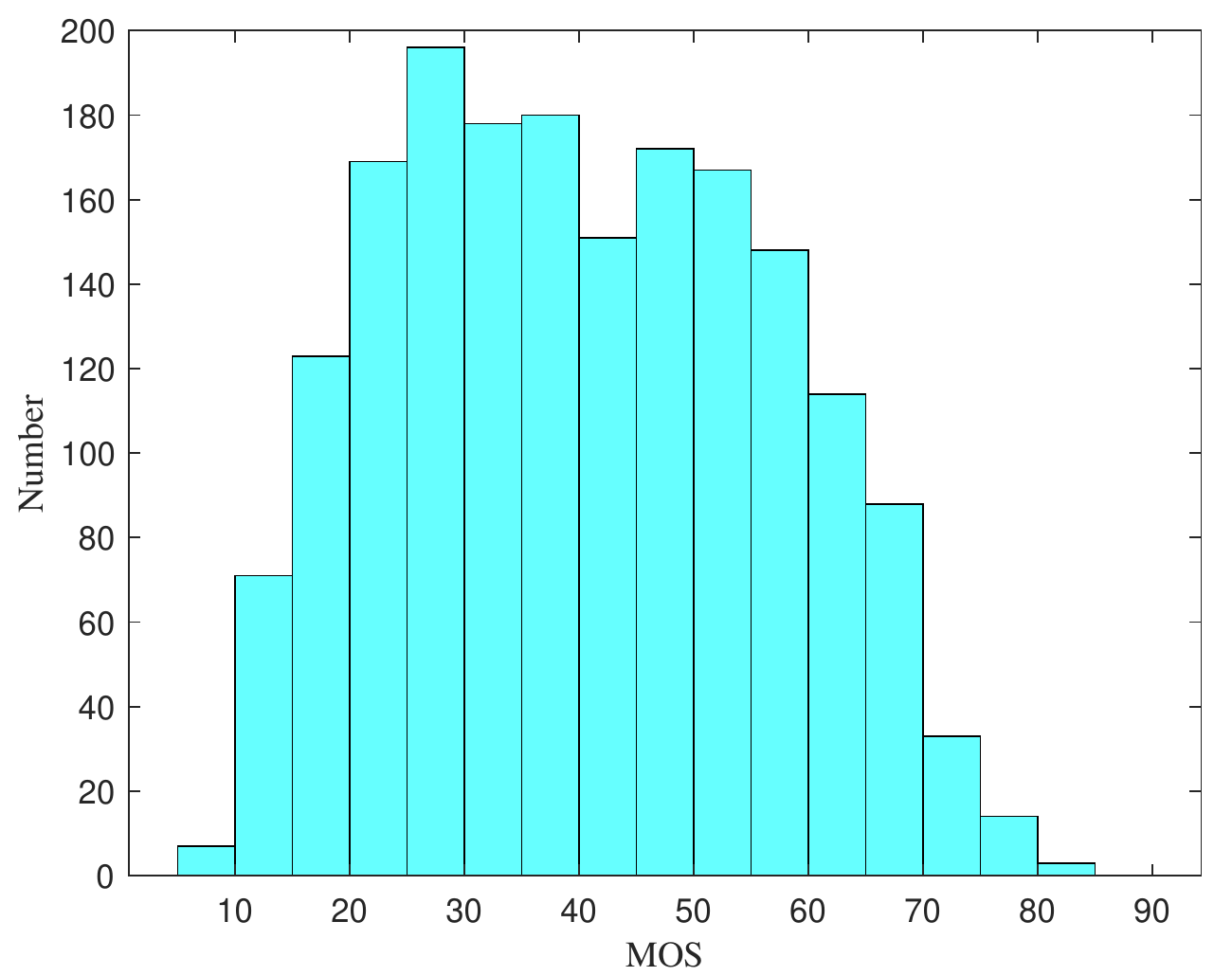}
    \caption{Enhanced videos}
    \label{h2}
  \end{subfigure}
  \caption{Detailed MOS distributions of the original and enhanced videos in our LLVE-QA dataset.}
  \label{histograms}
  \vspace{-2em}
\end{figure}

After the subjective experiment, we collect $45,320\enspace(\textit{i.e., } 22\times2,060)$ scores in total. Considering the standard deviation can well reflect the distribution of data, we calculate the standard deviation of 2,060 scores generated by each subject and reject two invalid subjects whose standard deviations of ratings are significantly lower than others. Finally, we obtain 20 valid subjects and MOSs for all videos in LLVE-QA dataset. In order to provide the insights in the difference between the MOSs of original videos and enhanced videos, we draw the MOS distributions in Figure \ref{histograms}.
The relatively uniform MOS distribution in Figure \ref{h2} reflects the diversity of visual quality of enhanced videos obtained by our selected LLVE algorithms.

\section{Proposed Method}
Benefiting from the built dataset, we propose a multi-dimensional quality assessment model named Light-VQA for low-light video enhancement. 
This model consists of the spatial and temporal information extraction module, the feature fusion module, and the quality regression module as shown in Figure \ref{architecture}. Specifically, spatial information extracted from key frames contains deep-learning-based semantic features, handcrafted brightness, and noise features. Temporal information extracted from video clips contains deep-learning-based motion features and handcrafted brightness consistency features. Then they are fused to obtain  the quality-aware representation. Finally, we utilize two Fully Connected (FC) layers to regress fused features into the video quality score.

\subsection{Spatial Information}

Since the adjacent frames of a video contain plenty of redundant contents, 
spatial information shows the extreme sensitivity to the video resolution and is not quite relevant to the video frame rate. Therefore, in order to reduce the computational complexity, we uniformly select $k$ key frames from the video to extract spatial information. 
In Light-VQA, we design two branches to simultaneously extract features in a video. Concretely, one is for deep-learning-based features, which contain rich semantic information, the other is for handcrafted features, which contain brightness and noise specifically designed for evaluating the quality of low-light and corresponding enhanced videos.

Swin Transformer~\cite{liu2021swin} has achieved more excellent performance than traditional CNNs in computer vision tasks such as image classification, object detection, and segmentation. For deep-learning-based features, we utilize the semantic information extracted from the last two stages of the pre-trained Swin Transformer:
\begin{equation}
   \begin{split}
    SF_{i} &= \alpha_{1} \oplus \alpha_{2},i \in\{1,\cdots,k\},  \\
    \alpha_{j} &= GAP(F_{i}^{j}), j \in\{1, 2\},
  \end{split}
\end{equation}
where $SF_{i}$ indicates the extracted semantic features of the $i$-th sampled key frame of a video, $\oplus$ represents the concatenation operation, $GAP(\cdot)$ stands for the global average pooling operation, $F_{i}^{j}$ indicates the feature maps in the $i$-th key frame generated from the $j$-th last stage of Swin Transformer, and $\alpha_{j}$ denotes the features after average pooling. For handcrafted features, we extract brightness and noise features which influence the quality of low-light and corresponding enhanced videos greatly~\cite{zhai2021perceptual} to better improve the quality-aware representation of Light-VQA:
\begin{equation}
    \begin{split}
    BF_{i} &= \Theta (F_{i}), \\
    NF_{i} &= \Psi (F_{i}),
    \end{split}
\end{equation}
where $BF_{i}$ and $NF_{i}$ indicate the extracted brightness and noise features from the $i$-th sampled key frame, respectively. $\Theta(\cdot)$ and $\Psi(\cdot)$ represent the extraction process of brightness and noise features, respectively. 

Therefore, given a video, we first uniformly select $k$ key frames, and then extract deep-learning-based and handcrafted features through two branches respectively. Finally, quality-aware spatial information is obtained by concatenating semantic features, brightness and noise features together:
\begin{equation}
    SI_{i} = SF_{i} \oplus BF_{i} \oplus NF_{i},
\end{equation}
where $SI_{i}$ indicates the ultimate spatial information of the $i$-th sampled key frame.

\begin{table*}[ht]
\begin{center}
\caption{ Experimental performance on our constructed LLVE-QA dataset and subset of KoNViD-1k. Our proposed Light-VQA achieves the best performance. ‘Handcrafted’ and ‘Deep Learning’ denote two types of leveraged features. The handcrafted models are inferior to deep-learning-based models. 
Best in \textbf{\textcolor{red}{red}} and second in \normalfont{\textcolor{blue}{blue}}.
}
\label{test}
\begin{tabular}{c|cc|ccc|ccc}
  \toprule[1.5pt]
 \multirow{2}{*}{VQA Model}  &   \multirow{2}{*}{Handcrafted} & \multirow{2}{*}{Deep Learning} & \multicolumn{3}{c|}{LLVE-QA} & \multicolumn{3}{c}{Subset of KoNViD-1k}\\ \cline{4-9}
 & & & SRCC$\uparrow$ & PLCC$\uparrow$ & RMSE$\downarrow$ & SRCC$\uparrow$ & PLCC$\uparrow$ &RMSE$\downarrow$ \\ \hline
V-BLIINDS (TIP, 2014)~\cite{saad2014blind}& \ding{52} &	 &	0.7123 &	0.7130 &	11.6185&	0.5927&	0.6157&	13.2098\\ 
TLVQM (TIP, 2019)~\cite{korhonen2019two}&\ding{52}	& 	&0.7321	& 0.7401 &	9.0957&	0.4260&	0.4671&	12.4164\\
VIDEVAL (TIP, 2021)~\cite{tu2021ugc}&\ding{52}& &	0.5294	&0.5233&	13.9555&	0.4963&	0.4052&	12.6448\\ 
RAPIQUE (OJSP, 2021)~\cite{tu2021rapique}&\ding{52}&\ding{52}&0.5890 & 0.5922 &	13.2555&	0.3861&	0.4751&	14.5661\\
Simple-VQA (ACM MM, 2022)~\cite{sun2022deep} & &\ding{52}&	0.8984	&0.8983& 7.2287&	0.6978&	0.7101&	\textcolor{blue}{10.0307}\\
FAST-VQA (ECCV, 2022)~\cite{wu2022fast}& &	\ding{52}&	\textcolor{blue}{0.9156} &\textcolor{blue}{0.9159}&	\textcolor{blue}{6.3528}&	\textcolor{blue}{0.7064}&	\textcolor{blue}{0.7156}&	10.7450\\ \hline
\textbf{Light-VQA} &\ding{52}&	\ding{52}&	\textbf{\textcolor{red}{0.9374}} & \textbf{\textcolor{red}{0.9393}}&	\textbf{\textcolor{red}{5.6523}}&	\textbf{\textcolor{red}{0.7975}}&	\textbf{\textcolor{red}{0.7860}}&	\textbf{\textcolor{red}{8.8070}}\\
\bottomrule[1.5pt]
\vspace{-1em}
\end{tabular}
\end{center}
\end{table*}

\subsection{Temporal Information}
Different from spatial information, temporal information is extremely susceptible to video frame-rate variations but not sensitive to resolution~\cite{sun2022deep}. Therefore, in order to preserve adequate temporal information while reducing computational complexity, we uniformly split the video into $k$ clips with lower resolution for temporal information extraction. Concretely, similar to the extraction of spatial information, we design two branches to obtain deep-learning-based and handcrafted features respectively. One is for motion, the other is for brightness consistency which is a significant feature in low-light videos.

For deep-learning-based features, we utilize a pre-trained SlowFast network~\cite{feichtenhofer2019slowfast} to extract motion features for each video clip:
\begin{equation}
    MF_{i} = \Phi (V_{i}),
\end{equation}
where $V_{i}$ indicates the $i$-th video clip, $\Phi(\cdot)$ denotes the extraction operation of motion features, and $MF_{i}$ stands for the extracted motion features from the $i$-th video clip. For handcrafted features, we extract brightness consistency features:
\begin{equation}
    CF_{i} = \Gamma (V_{i}),
\end{equation}
where $\Gamma$ indicates the extraction operation of brightness consistency features, and $CF_{i}$ denotes the extracted brightness consistency features from the $i$-th video clip. 

To sum up, given a video, we split it into $k$ clips with lower resolution uniformly, and then extract deep-learning-based and handcrafted features through two branches respectively. Finally, temporal information is obtained by concatenating motion features and brightness consistency features together:
\begin{equation}
    TI_{i} = MF_{i} \oplus CF_{i},
\end{equation}
where $TI_{i}$ indicates the temporal information of the the $i$-th video clip.

\subsection{Spatial-Temporal Fusion}

After obtaining the both spatial and temporal information, it is essential to fuse them to get a more comprehensive feature expression. In this paper, we utilize Multi-Layer Perception (MLP) as the fusion module to integrate spatial with temporal information due to its simplicity and effectiveness. Various fusion strategies based on attention mechanisms can be included, but they are beyond the scope of this paper.
Specifically, given spatial information $SI_{i}$ extracted from the $i$-th key frame and temporal information $TI_{i}$ extracted from the $i$-th video clip, we concatenate them first and then pass them through a MLP:
\begin{equation}
    FF_{i} = \mathcal{F}(SI_{i} \oplus TI_{i}),
\end{equation}
where $\mathcal{F}(\cdot)$ indicates the learnable feature fusion that contains one FC layer with 1,024 neurons and one REctified Linear Unit (RELU), and $FF_{i}$ represents features after fusion of the $i$-th video clip (the $i$-th key frame can be regarded as one frame in the $i$-th video clip but with original resolution). All elements in $FF_{i}$ are calculated jointly by $SI_{i}$ and $TI_{i}$.

\subsection{Quality Regression}
Subsequently, we utilize another two FC layers to regress quality-aware representation $FF_{i}$ into the video quality score:
\begin{equation}
    Q_{i} = FC(FF_{i}), 
\end{equation}
where $Q_{i}$ indicates the quality score of the $i$-th video clip. Finally, the overall score of the entire video is obtained by averaging the quality scores of all $k$ video clips:
\begin{equation}
    Q = \frac{1}{k}\sum_{i=1}^{k}Q_{i},
\end{equation}
where $Q$ is the quality score of the video and $k$ indicates the number of video clips.

Our loss function for training is composed of two parts~\cite{sun2022deep}: Mean Absolute Error (MAE) loss and rank loss~\cite{wen2021strong}. MAE loss is widely used in various deep learning tasks, and in this paper it is defined as:
\begin{equation}
    L_{MAE} = \frac{1}{N}\sum_{m=1}^{N}|Q_{m} - \hat{Q}_{m}|,
\end{equation}
where $\hat{Q}_{m}$ is the predicted MOS for the $m$-th video in a batch and N is the batch size. Rank loss can help the network to learn the relative quality of different videos, which exactly coincides with our need to compare the quality of different LLVE algorithms. Specifically, the rank loss is defined as follows:
\begin{equation}
    L_{rank} = \frac{1}{N^2}\sum_{m=1}^{N}\sum_{n=1}^{N} max(0, |\hat{Q}_{m} -\hat{Q}_{n}| - e(\hat{Q}_{m}, \hat{Q}_{n})\cdot(Q_{m} - Q_{n})),
\end{equation}
where $m$ and $n$ are two different videos in one training batch, and $e(\hat{Q}_{m}, \hat{Q}_{n})$ is formulated as:
\begin{equation}
    e(\hat{Q}_{m}, \hat{Q}_{n}) = 
    \left\{ 
    \begin{array}{lc}
     1, \quad &\hat{Q}_{m} \geq \hat{Q}_{n}, \\
    -1, \quad &\hat{Q}_{m} \textless \hat{Q}_{n},
    \end{array}
    \right.
\end{equation}
Then, the optimization objective can be obtained by:
\begin{equation}
    L = L_{MAE} + \beta \cdot L_{rank},
\end{equation}
where $\beta$ is a hyper-parameter for balancing the MAE loss and the rank loss.

\section{Experiment}


\begin{figure*}[ht]
\centering
\begin{subfigure}{0.24\textwidth}
    \includegraphics[width=\textwidth]{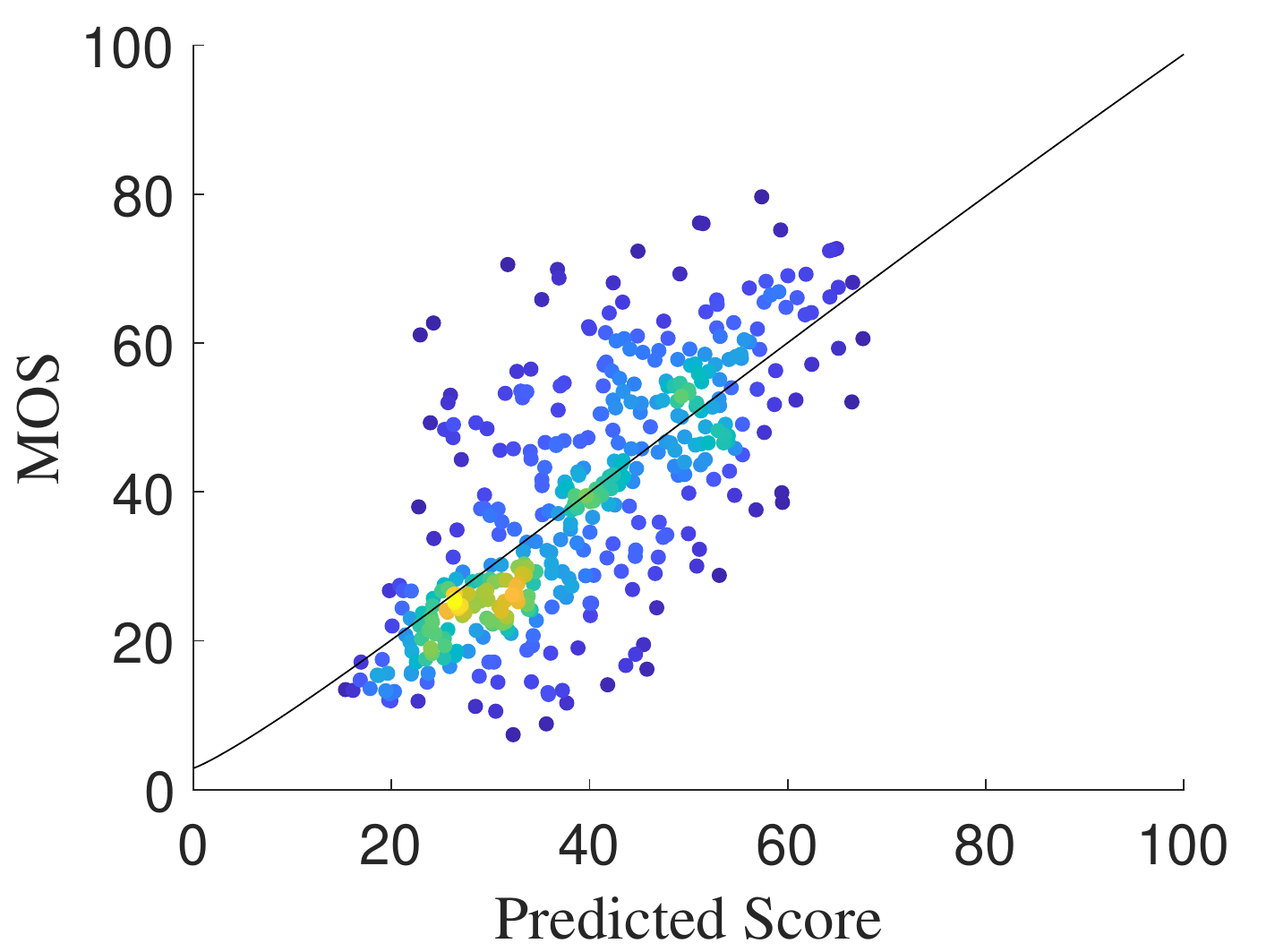}
    \caption{V-BLIINDS}
    \label{fig:1}
\end{subfigure}
\begin{subfigure}{0.24\textwidth}
    \includegraphics[width=\textwidth]{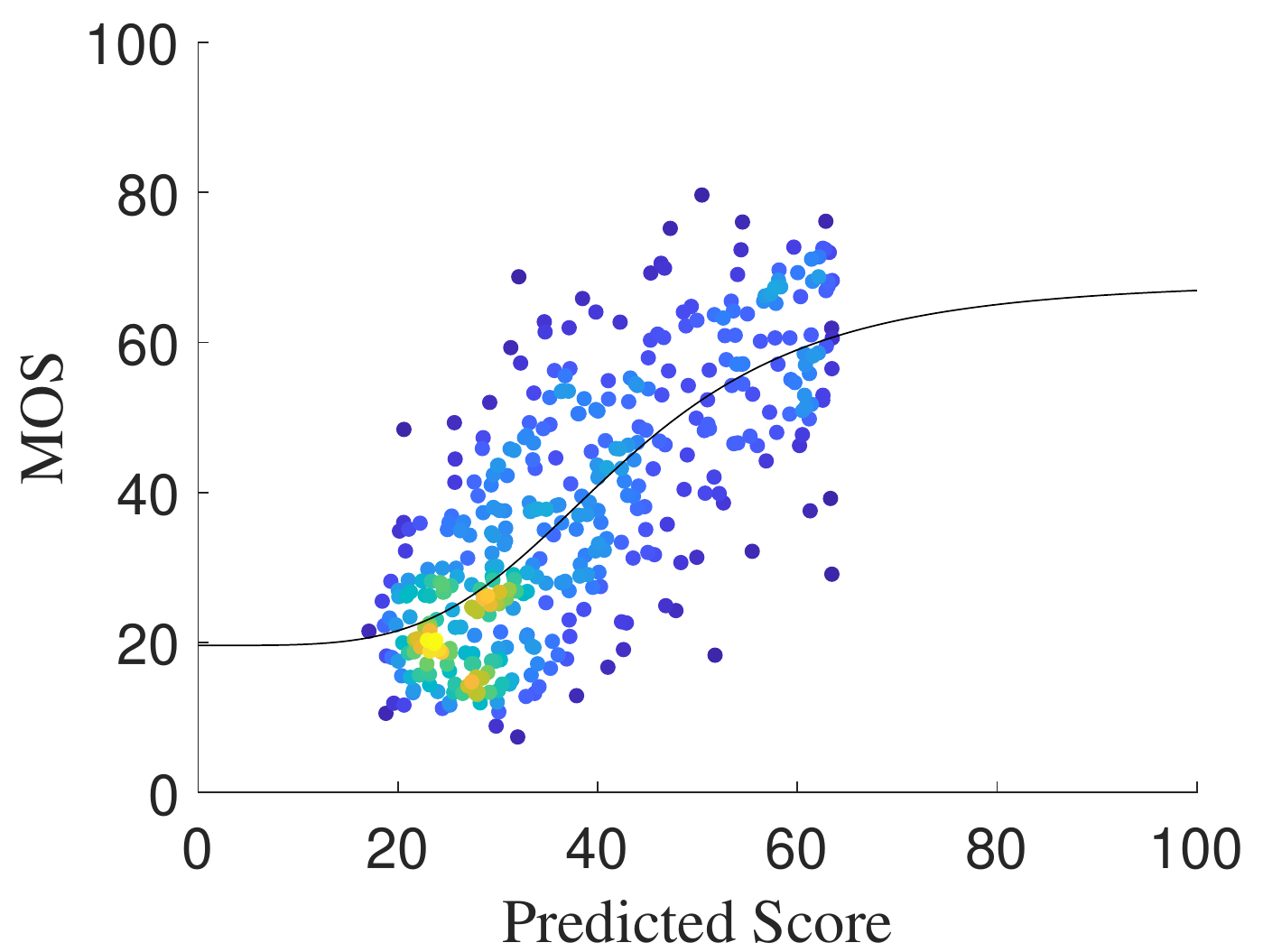}
    \caption{TLVQM}
    \label{fig:2}
\end{subfigure}
\begin{subfigure}{0.24\textwidth}
    \includegraphics[width=\textwidth]{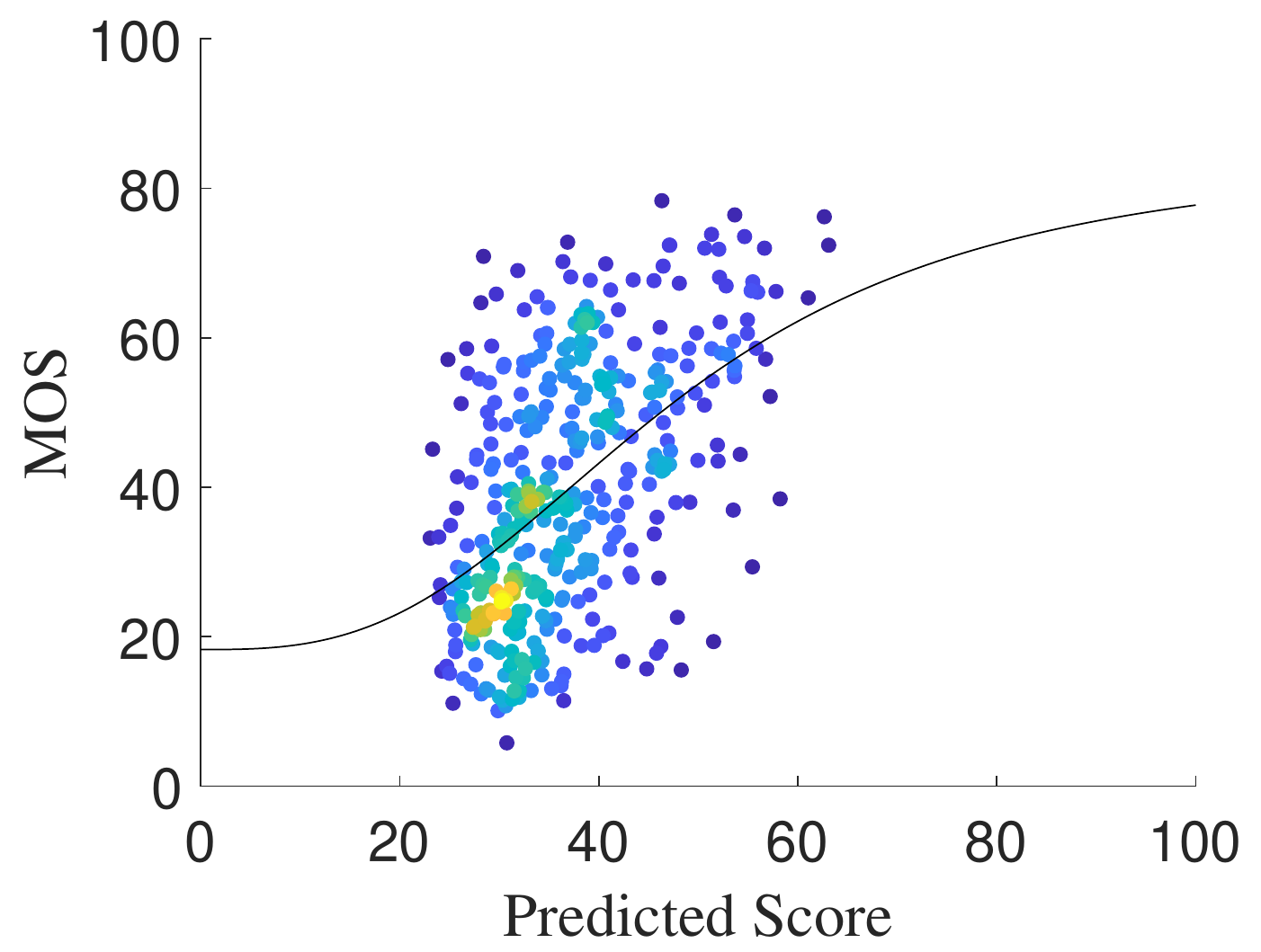}
    \caption{VIDEVAL}
    \label{fig:3}
\end{subfigure}
\begin{subfigure}{0.24\textwidth}
    \includegraphics[width=\textwidth]{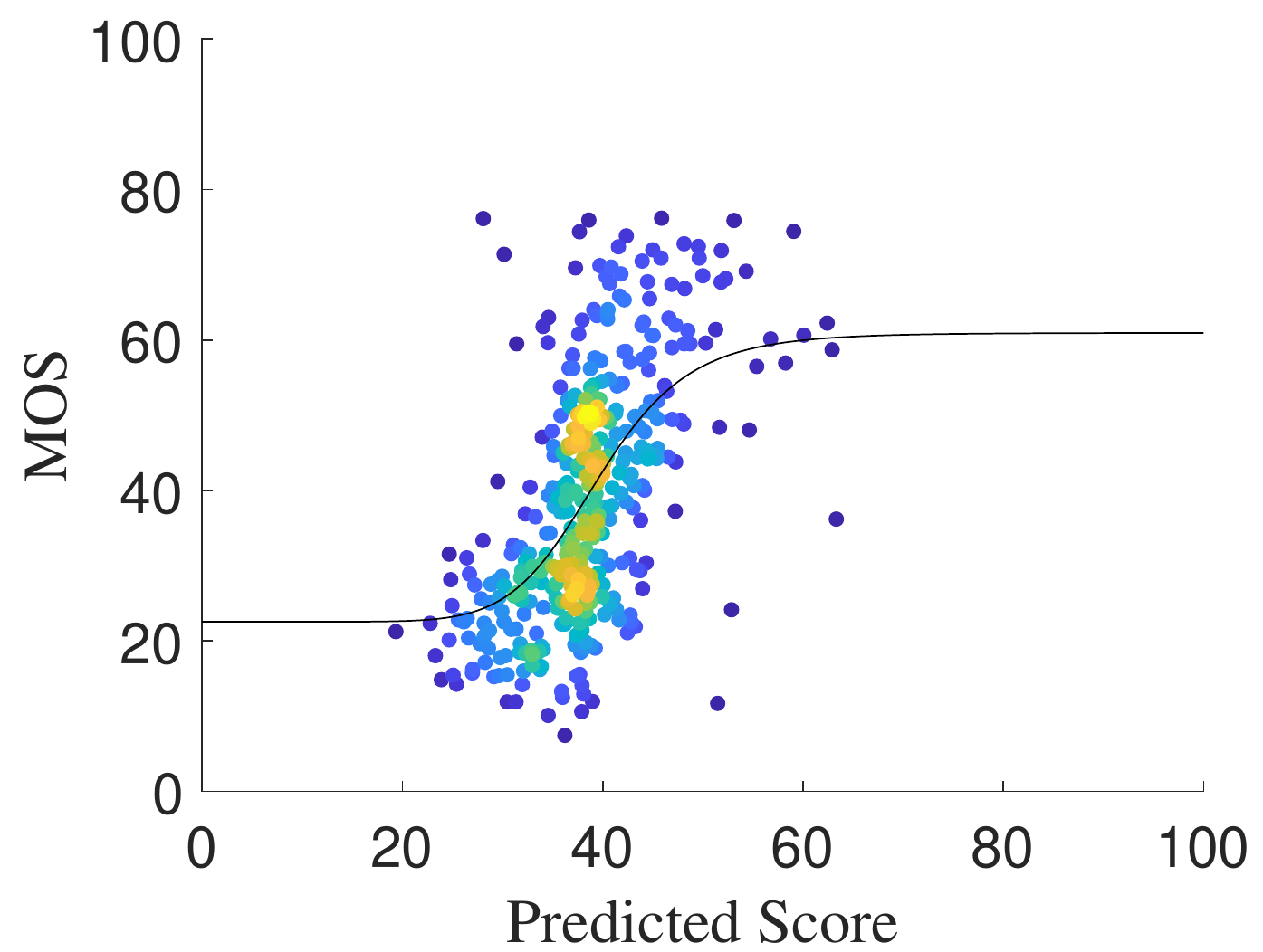}
    \caption{RAPIQUE}
    \label{fig:6}
\end{subfigure}
\begin{subfigure}{0.24\textwidth}
    \includegraphics[width=\textwidth]{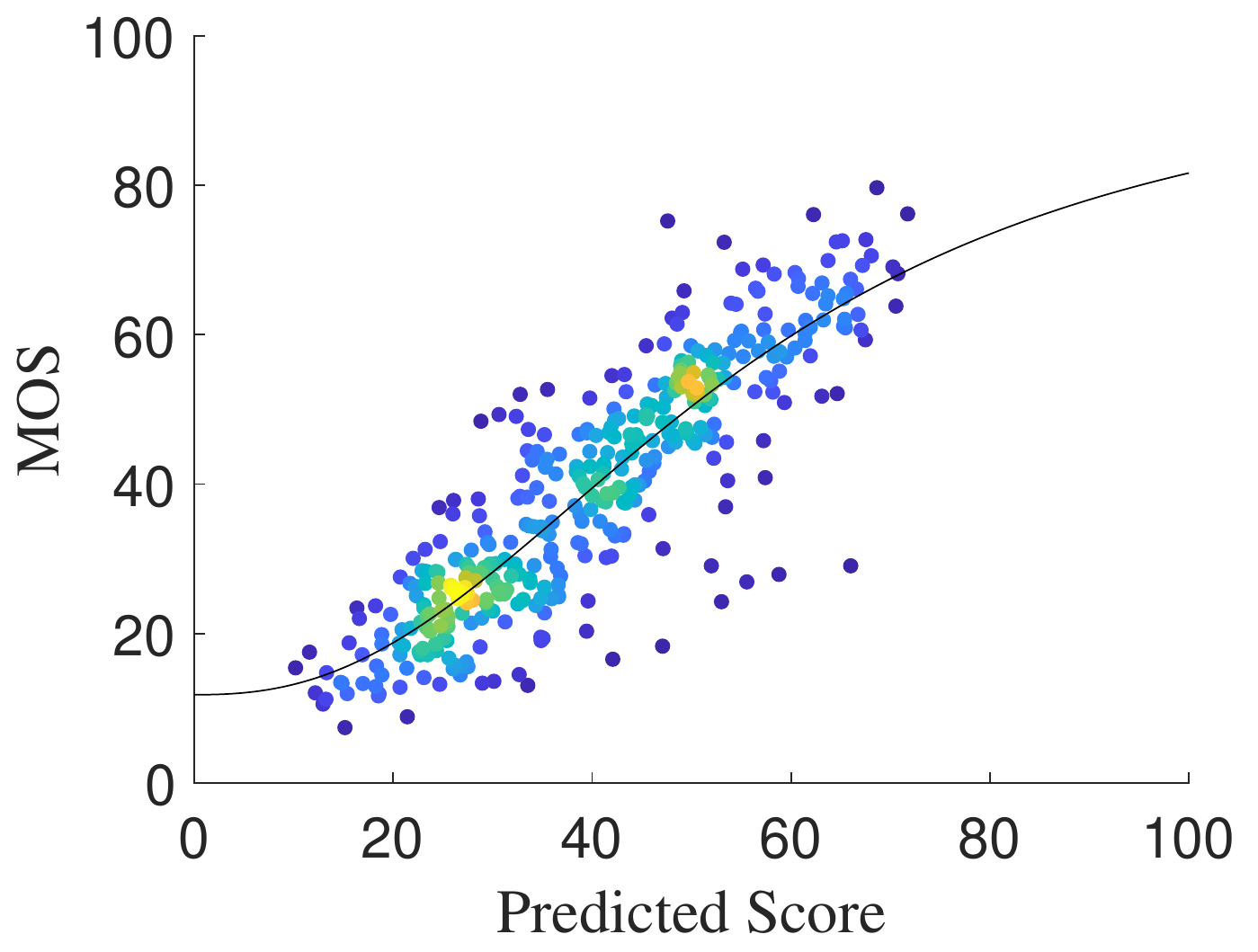}
    \caption{Simple-VQA}
    \label{fig:4}
\end{subfigure}
\begin{subfigure}{0.24\textwidth}
    \includegraphics[width=\textwidth]{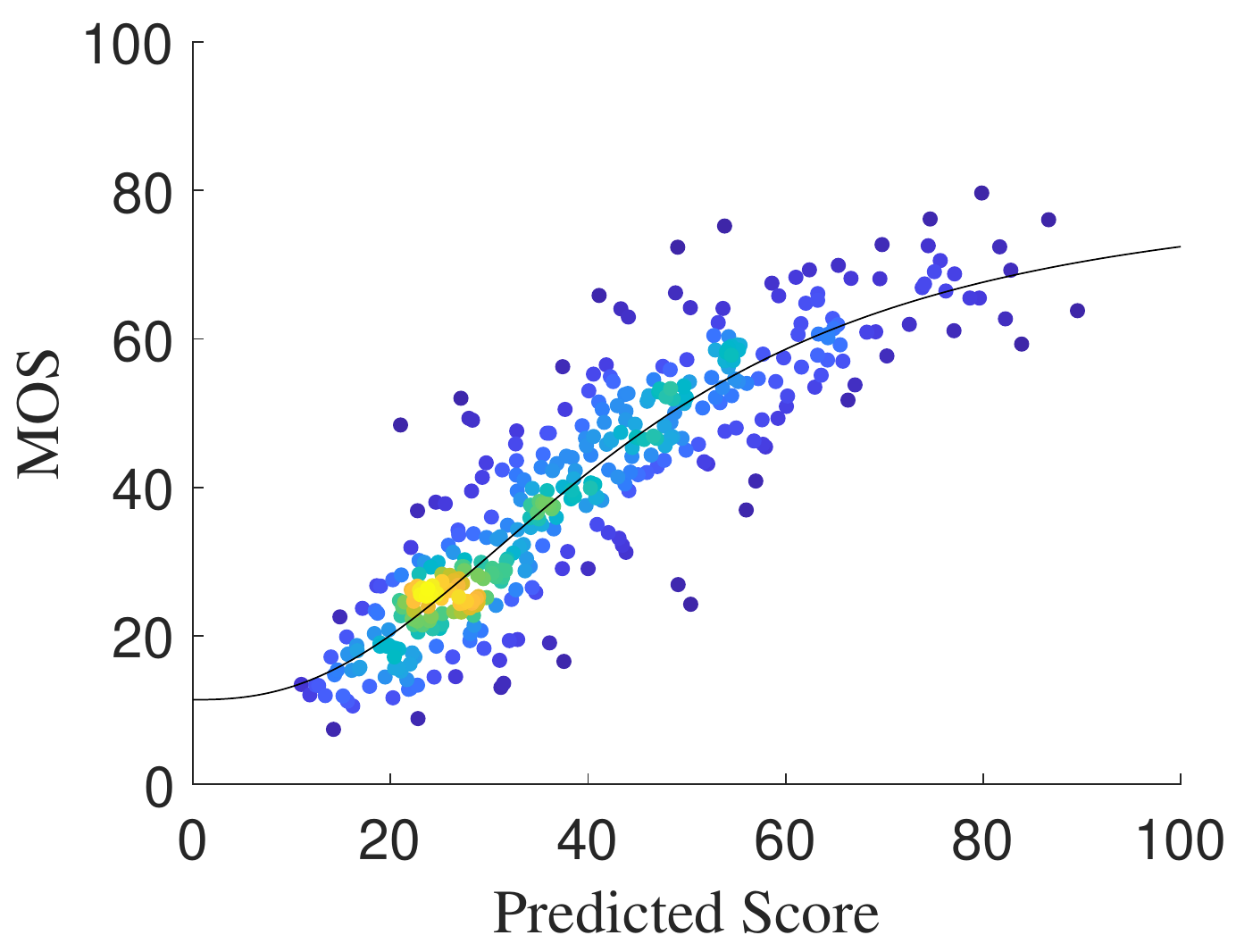}
    \caption{FAST-VQA}
    \label{fig:5}
\end{subfigure}
\begin{subfigure}{0.24\textwidth}
    \includegraphics[width=\textwidth]{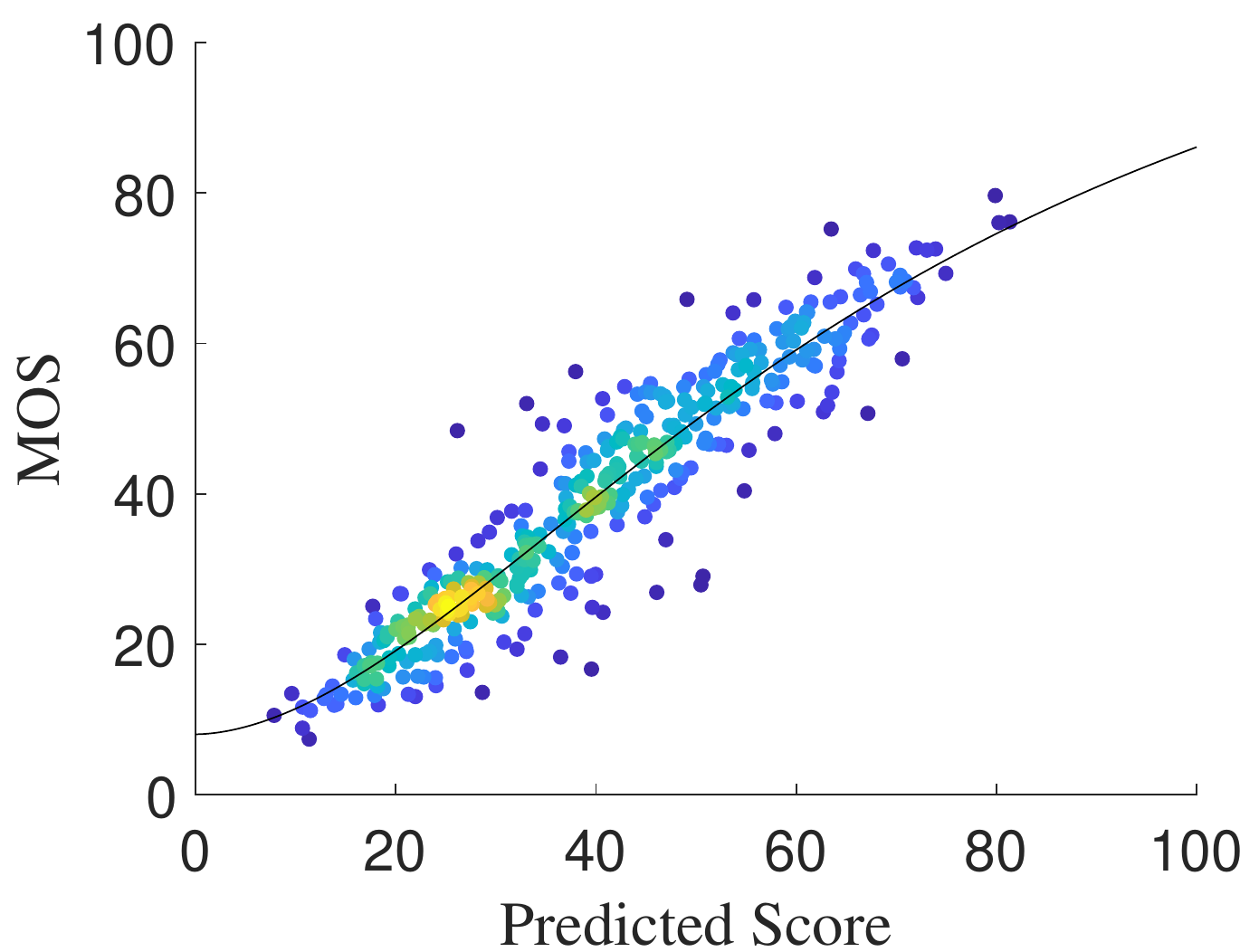}
    \caption{Light-VQA}
    \label{fig:7}
\end{subfigure}
\caption{The scatter plots of the predicted scores versus the MOSs. The curves are obtained by a four-order polynomial nonlinear fitting. It is evident that the predicted scores of our proposed VQA bear the closest resemblance to the MOSs.}
\label{Scatter plots}
\vspace{-1em}
\end{figure*}



\subsection{Performance Comparisons with the SOTA VQA Models}
To validate the effectiveness of Light-VQA on LLVE-QA dataset, we compare it with $6$ state-of-the-art VQA models including V-BLIINDS~\cite{saad2014blind}, TLVQM~\cite{korhonen2019two}, VIDEVAL~\cite{tu2021ugc}, RAPIQUE~\cite{tu2021rapique}, Simple-VQA~\cite{sun2022deep}, and FAST-VQA~\cite{wu2022fast}. 
We utilize the same training strategy to train all models on the LLVE-QA dataset and ensure their convergence. Then we test them on the testing set. The numbers of videos in training set, validation set, testing set are 1260, 400, and 400 respectively. 
The overall experimental results are shown in Table \ref{test}.
Figure \ref{Scatter plots} shows the scatter plots of the predicted MOSs versus the ground-truth MOSs on LLVE-QA dataset for $7$ VQA models listed in Table \ref{test}. The curves shown in Figure \ref{Scatter plots} are obtained by a four-order polynomial nonlinear fitting. According to Table \ref{test}, Light-VQA achieves the best performance in all $7$ models and leads the second place (\textit{i.e.}, FAST-VQA) by a relatively large margin, which demonstrates its effectiveness for the perceptual quality assessment of low-light video enhancement.


\begin{table*}[t]
\begin{center}
\caption{ Experimental results of ablation studies on LLVE-QA dataset.  Best in \textbf{\textcolor{red}{red}} and second in \normalfont{\textcolor{blue}{blue}}. [Key: SF: Semantic Features, BF: Brightness Features, NF: Noise Features, MF: Motion Features, CF: Brightness Consistency Features, FF: Feature Fusion, MLR: Multiple Linear Regression]}
\label{ablation}
\begin{tabular}{c|cc|cc|cc|ccc}
  \toprule[1.5pt]
\multirow{2}{*}{Model}  & \multicolumn{2}{c|}{Spatial Information}  & \multicolumn{2}{c|}{Temporal Information}  & \multicolumn{2}{c|}{Fusion Method}  &  \multicolumn{3}{c}{LLVE-QA}  \\ \cline{2-10}
&SF &BF + NF &MF &CF & FF&MLR & SRCC$\uparrow$ & PLCC$\uparrow$ & RMSE$\downarrow$ \\ \hline
  1    &\ding{52}  &	     &	  &	   &    &     & 0.9120     & 0.9143 & 6.5377  \\  
  2    &	         & 	&\ding{52} &    &    &     & 0.8446     & 0.8438  & 8.8438  \\  
  3    &\ding{52}  & 	&\ding{52} &    &\ding{52}    &     & 0.9223     & 0.9245  & 6.6829  \\  
  4    &\ding{52}  &\ding{52} 	&\ding{52} &    &\ding{52}    &     & \textcolor{blue}{0.9324}     & \textcolor{blue}{0.9354}  & \textcolor{blue}{5.8278}  \\  
  5    &\ding{52}  & 	&\ding{52} & \ding{52}  &\ding{52}    &     & 0.9299     & 0.9310   & 5.9764 \\  
  6    &\ding{52}  &\ding{52} 	&\ding{52} & \ding{52}  &    &\ding{52}     & 0.9231     & 0.9243   & 6.7132 \\  
  7    &\ding{52}  &\ding{52} 	&\ding{52} & \ding{52}  &\ding{52}    &     & \textbf{\textcolor{red}{0.9374}}     & \textbf{\textcolor{red}{0.9393}}  & \textbf{\textcolor{red}{5.6523}} \\  \bottomrule[1.5pt]
\end{tabular}
\end{center}
\end{table*}

\subsection{Cross Dataset Performance}

\begin{figure}[t]
\setlength{\abovecaptionskip}{0cm} 
\setlength{\belowcaptionskip}{0.1cm}
  \centering
  \begin{subfigure}[b]{0.49\linewidth}
    \centering
    \includegraphics[width=\linewidth]{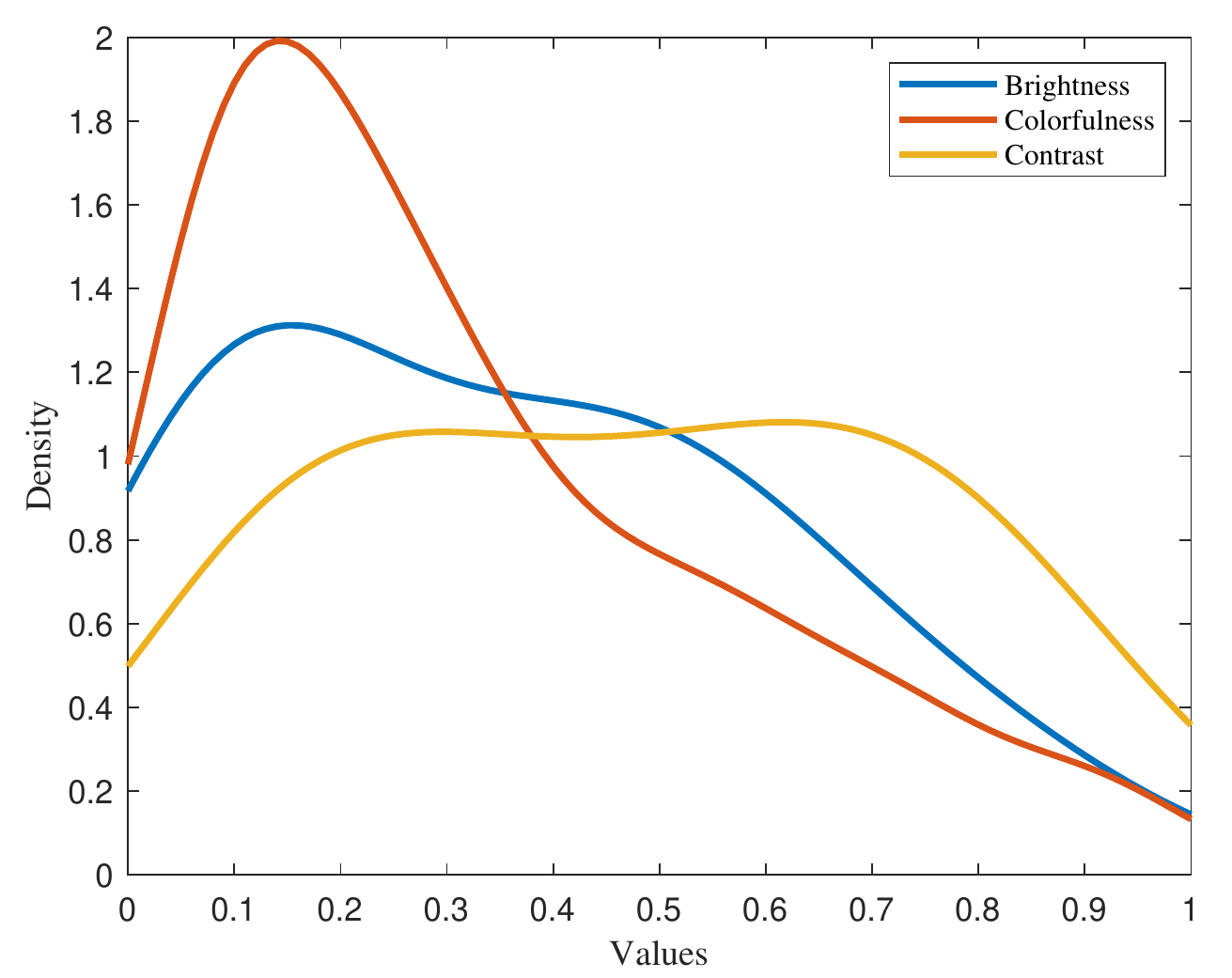}
    \caption{Attribute values}
    \label{llv1}
  \end{subfigure}
  \hfill
  \begin{subfigure}[b]{0.49\linewidth}
    \centering
    \includegraphics[width=\linewidth]{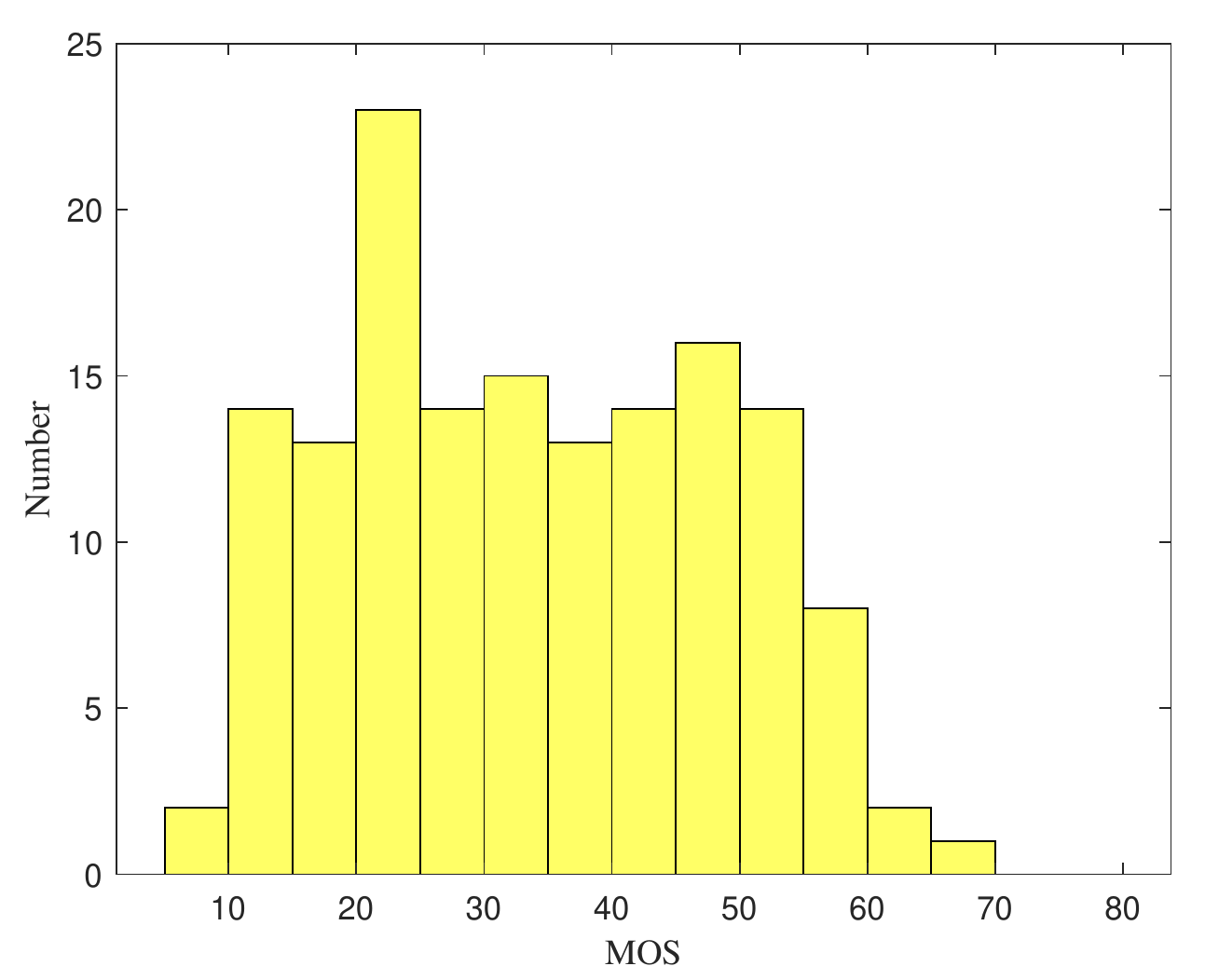}
    \caption{MOS}
    \label{llv2}
  \end{subfigure}
  \caption{Detailed attribute and MOS distribution for low light video subset of KoNViD-1k~\cite{hosu2017konstanz}.}
  \label{llv}
  \vspace{-2.2em}
\end{figure}

To examine the cross-dataset performance of the model, we conduct experiments on the subset of low light videos in KoNViD-1k~\cite{hosu2017konstanz}. The distributions of three attributes (\textit{i.e.}, brightness, colorfulness and contrast) and MOS of the subset are shown in Figure \ref{llv}. We directly leverage the models pre-trained on LLVE-QA dataset to perform testing on the newly built subset with ease and efficiency. The overall experimental results on subset of KoNViD-1k are shown in Table \ref{test}. Since LLVE-QA dataset includes both low-light videos and their corresponding enhanced versions, whereas KoNViD-1k exclusively consists of low-light videos, the quality-aware representation learned from LLVE-QA dataset is less effective on KoNViD-1k. All methods have experienced the decline of performance. However, our proposed Light-VQA still surpasses the other 6 VQA methods by a large margin, which demonstrates its good generalization ability in terms of the quality assessment of low light videos. 

\subsection{Ablation Studies}
 In this subsection, a series of ablation experiments are conducted to analyze the contributions of different modules in Light-VQA. Table \ref{ablation} shows the experimental results of ablation studies. \textit{Model 1} only utilizes Semantic Features (SF) extracted by Swin Transformer~\cite{liu2021swin}. \textit{Model 2} only utilizes Motion Features (MF) extracted by SlowFast R50~\cite{feichtenhofer2019slowfast}. \textit{Model 3} utilizes both SF and MF, and obtains the results after passing them through the Feature Fusion (FF) module. Based on \textit{Model 3}, \textit{Model 4} adds handcrafted Brightness and Noise Features (BF + NF) that belong to spatial information together with SF. Based on \textit{Model 3}, \textit{Model 5} adds handcrafted Brightness Consistency Features (CF) that belong to temporal information coupled with MF. \textit{Model 6} utilizes all the spatial information and temporal information, but instead of performing feature fusion, Multiple Linear Regression (MLR) is used as a replacement. \textit{Model 7} is the complete model we propose, in which we fuse all the spatial and temporal information, and obtain the best results.

\textbf{Feature Extraction Module.} For Light-VQA, both spatial and temporal information is composed of deep-learning-based and handcrafted features. First, \textit{Model 1} and \textit{Model 2} are designed to verify the contribution of deep-learning-based features in spatial information and temporal information, respectively. It can be observed from the results that semantic features in spatial information are significantly superior to motion features in temporal information. When we fuse them in \textit{Model 3}, the performance of the model is further improved. Second, based on \textit{Model 3}, \textit{Model 4}, and \textit{Model 5} are designed to prove the effectiveness of handcrafted features in spatial information and temporal information respectively. It is evident that both of them obtain better results compared to \textit{Model 3}. When we add them all in \textit{Model 7}, the final model Light-VQA exhibits the best performance.

\textbf{Feature Fusion Module.} In this paper, we utilize MLP as the feature fusion module to integrate spatial-temporal information. To verify its effectiveness, we train two models separately, one of which only contains temporal information and the other only contains spatial information, and then we utilize Multiple Linear Regression (MLR) to get the predicted score:
\begin{equation}
    Q_{i}^{m} = a\cdot Q_{i}^{s} + b\cdot Q_{i}^{t} + c,
\end{equation}
where $Q_{i}^{s}$ indicates the score obtained by spatial information module, $Q_{i}^{t}$ indicates the score obtained by temporal information module, and $Q_{i}^{m}$ denotes the score after MLR. $a, b$, and $c$ are parameters to be fitted in MLR. By comparing the results of \textit{Model 6} and \textit{Model 7} in Table \ref{ablation}, it is evident that our feature fusion module plays a role in improving the prediction performance.

\begin{table}
\centering
\caption{The average scores predicted by Light-VQA on original low-light videos, results of StableLLVE w/o Light-VQA, and results of StableLLVE w/ Light-VQA.}
\label{refinement}
\begin{tabular}{c|ccc}
\toprule[1.5pt]
 Dataset & Origin & w/o Light-VQA & w/ Light-VQA \\
\hline
Average score&  39.0933 &  59.4926 &  \textbf{86.2688}  \\
\bottomrule[1.5pt]
\end{tabular}
\vspace{-1.5em}
\end{table}

\subsection{Refinement for Training LLVE Algorithms}
\begin{figure}[ht]
\centering
\begin{subfigure}[t]{0.15\textwidth}
    \includegraphics[width=1\textwidth]{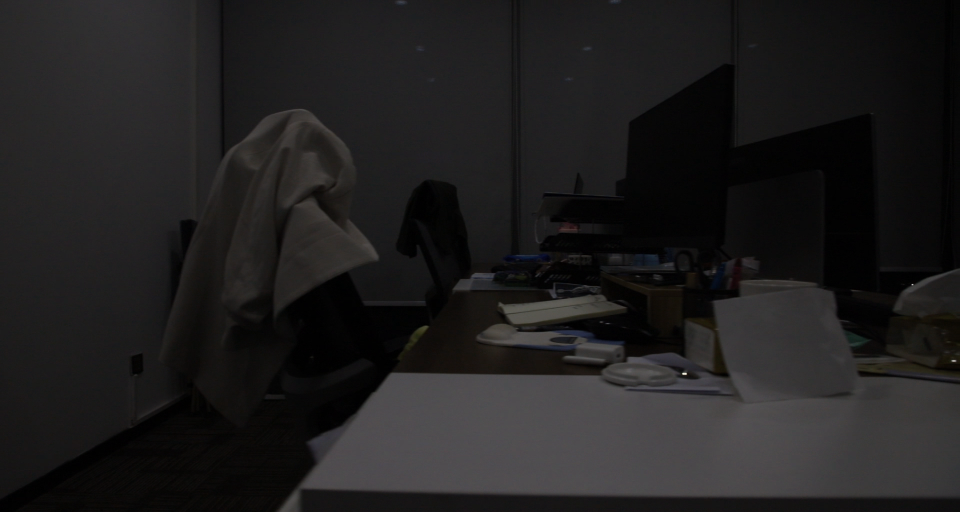}
    \subcaption*{(a)\ Origin}
\end{subfigure}
\begin{subfigure}[t]{0.15\textwidth}
    \includegraphics[width=1\textwidth]{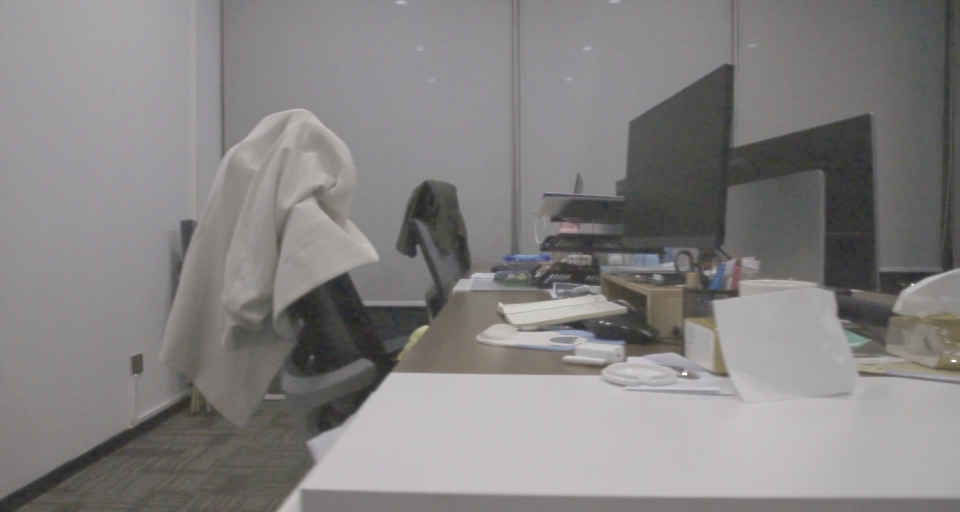}
    \subcaption*{(b)\ w/o Light-VQA}
\end{subfigure}
\begin{subfigure}[t]{0.15\textwidth}
    \includegraphics[width=1\textwidth]{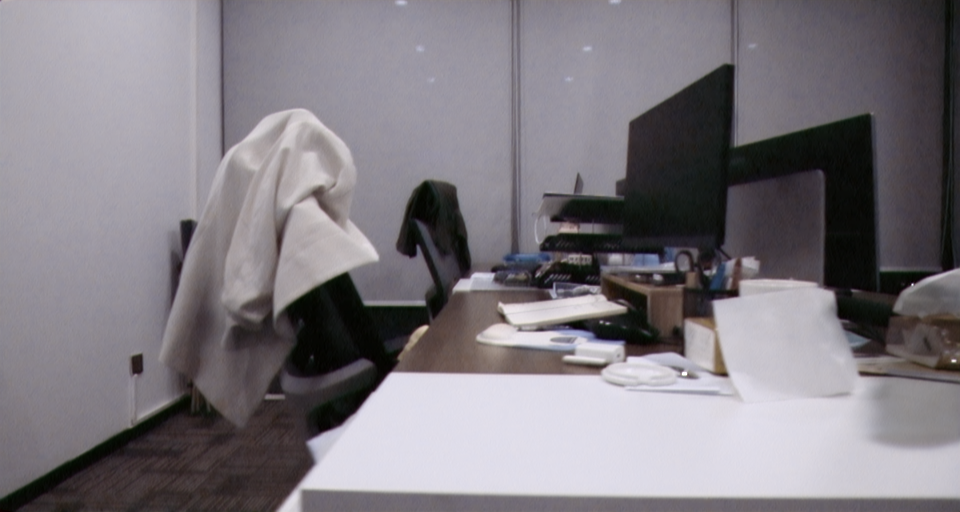}
    \subcaption*{(c)\ w/ Light-VQA}
\end{subfigure}
\caption{The detailed qualitative comparisons of original low-light videos, results of StableLLVE w/o Light-VQA, and results of StableLLVE w/ Light-VQA.}
\label{sample}
\vspace{-1em}
\end{figure}

To demonstrate that Light-VQA can be utilized as a metric to facilitate the development of LLVE algorithms by approaching the human visual system, we use Light-VQA as a loss function to train a recent low-light video enhancement algorithm named StableLLVE~\cite{zhang2021learning}. Experimental results show that training with Light-VQA as a loss function yields videos with better perceptual quality compared to the original training method. 
The dataset we use for experiments is from SDSD~\cite{wang2021seeing}. The average scores predicted by Light-VQA on original low-light videos, results of StableLLVE w/o Light-VQA, and results of StableLLVE w/ Light-VQA are shown in Table \ref{refinement}. The detailed qualitative comparisons of are shown in Figure \ref{sample}. It is evident that the results of StableLLVE training with Light-VQA have better perceptual quality.

\section{Conclusion}
In this paper, we focus on the issue of evaluating the quality of LLVE algorithms. To facilitate our work, we construct a LLVE-QA dataset containing 2,060 videos. Concretely, we collect 254 original low-light videos that contain various scenes and generate the remaining videos by utilizing different LLVE algorithms. Further, we propose an effective VQA model named Light-VQA specially for low-light video enhancement. Concretely, we integrates the luminance-sensitive handcrafted features into deep-learning-based features in both spatial and temporal information extractions. Then we fuse them to obtain the overall quality-aware representation.
Extensive experimental results have validated the effectiveness of our Light-VQA. For future work, we will enable the Light-VQA to evaluate the recovery performance of overexposed videos.

\begin{acks}
This work was supported in part by the Shanghai Pujiang Program under Grant 22PJ1406800, in part by the National Natural Science Foundation of China under Grant 62225112 and Grant 61831015.
\end{acks}

\bibliographystyle{ACM-Reference-Format}
\balance
\bibliography{sample-base}


\end{document}